\documentclass[final,onefignum,onetabnum]{siamonline250211}

\usepackage{smile_siam}
\usepackage{xcolor}
\usepackage{float}
\usepackage{wrapfig}
\newsiamthm{setting}{Setting}
\newsiamthm{scenario}{Scenario}
\newtheorem{remark}{Remark}

\title{
A Geometric Understanding of Denoising Autoencoder with Applications to Image Denoising}

\author{Haoran Yan\thanks{School of Mathematics, Georgia Institute of Technology, Atlanta, GA, USA.}
\and
Rongjie Lai\thanks{Department of Mathematics, Purdue University, West Lafayette, IN, USA.}
\and
Wenjing Liao\thanks{School of Mathematics, Georgia Institute of Technology, Atlanta, GA, USA.}
}
\begin{document}

\maketitle

\begin{abstract}
Denoising autoencoders have shown strong empirical performance in recovering clean data from their noisy observations, yet their underlying statistical behavior remains underexplored. In this paper, we study denoising autoencoders under the manifold hypothesis, assuming that clean data are supported on low-dimensional smooth manifolds embedded in a high-dimensional ambient space, while noisy observations are generated by isotropic Gaussian noise. We establish generalization guarantees in three geometric scenarios: a single manifold, a pairwise separated finite union of manifolds, and a finite union of intersecting manifolds. Using local chart representations, partitions of unity, and neural network approximation, we derive upper bounds for the expected squared reconstruction error for denoising autoencoder. In the single-manifold and separated-union scenarios, the error consists of a sample-dependent statistical term of order $n^{-2/(d+2)}\log^4 n$, a residual term of order $d\sigma^2/D$ caused by tangential noise, and an exponentially small tail term associated with failure of the manifold projection. Here $D$ is the data ambient dimension, $d$ is the intrinsic dimension of the clean-data manifold, and $\sigma^2$ is the expected squared noise magnitude. In the scenario of intersecting manifolds, we further identify an additional ambiguity term determined by the probability mass near the intersection neighborhood. Numerical experiments on synthetic manifolds and real image datasets support the predicted sample-size scaling law and illustrate the additional reconstruction ambiguity near manifold intersections.
\end{abstract}

\begin{keywords}
denoising autoencoder, image denoising, manifold hypothesis, generalization error, neural network approximation
\end{keywords}

\section{Introduction}

High-dimensional data arise in many modern machine learning problems, including image analysis, scientific computing, and biomedical imaging. Deep neural networks have achieved strong empirical performance in learning representations of such high-dimensional data. Among the many architectures used in deep learning, autoencoders have emerged as one of the most popular tools for nonlinear representation learning and dimension reduction \cite{bourlard1988auto,kramer1991nonlinear,hinton1993autoencoders,liou2014autoencoder}. A standard autoencoder consists of an encoder that maps an input to a latent representation and a decoder that maps the latent representation back to the data space. By training the encoder and decoder jointly through a reconstruction objective, autoencoders can capture features and nonlinear structures in data.

Denoising autoencoders extend this idea by training the network to reconstruct clean data from noisy observations \cite{vincent2008extracting,bengio2009learning}. Given a clean sample $\xb$, a noisy observation $\yb$ is generated by adding a perturbation to $\xb$. The denoising autoencoder takes $\yb$ as input and uses $\xb$ as the reconstruction target. This training mechanism encourages the learned maps to remove noise while preserving the underlying data structure, and has been widely used to learn robust representations from noisy data. 

Despite their empirical success, the statistical behavior of denoising autoencoders is not yet well understood, particularly in high-dimensional settings. In high ambient dimensions, statistical estimation problems can in principle suffer from the curse of dimensionality. However, many high-dimensional datasets are believed to possess much lower-dimensional intrinsic structure, as supported by empirical evidence in many image and language datasets \cite{tenenbaum290global,roweis2000nonlinear,coifman2005geometric,pope2020intrinsic,havrilla2024understanding}. The manifold hypothesis provides a natural geometric framework for describing such structure by modeling clean data as being supported on or near a $d$-dimensional manifold embedded in a high-dimensional Euclidean space in $\RR^D$ with $d\ll D$ \cite{tenenbaum290global,roweis2000nonlinear,fefferman2016testing}. This viewpoint raises a fundamental question for denoising autoencoders: how does the reconstruction error decrease with the number of noisy and clean training pairs, and is this sample-size scaling law governed primarily by the intrinsic dimension $d$ or by the ambient dimension $D$?

The manifold hypothesis has motivated a growing body of theory showing that neural networks can adapt to low-dimensional geometric structure, with approximation and statistical rates depending primarily on the intrinsic dimension rather than the ambient dimension \cite{shaham2018provable,chen2019efficient,chen2019nonparametric,cloninger2021deep,liu2022benefits,liu2022adaptive,nakada2020adaptive}. A theoretical line of work has studied denoising autoencoders through their connection with score matching. It is shown in \cite{vincent2011connection}  that, under Gaussian corruption, the denoising autoencoder objective admits a population-level interpretation related to the score of the corrupted data distribution. However, this perspective does not directly yield finite-sample generalization guarantees for denoising autoencoders under a low-dimensional manifold model. A closely related line of work studies chart autoencoders for manifold-structured data. By using multiple local coordinate charts rather than a single global latent parametrization, chart autoencoders can represent manifolds with nontrivial topology \cite{schonsheck2019chart}. \cite{liu2024chart} developed approximation and generalization theories for such models and established rates governed mainly by the intrinsic dimension. Their analysis provides much of the geometric and statistical framework used in our work, but does not consider the isotropic Gaussian denoising setting or the separated and intersecting multi-manifold settings considered here.
 
In this paper, we study the statistical performance of denoising autoencoders trained from noisy and clean sample pairs under the manifold hypothesis. We consider a common data setting in which clean samples are drawn from low-dimensional structures embedded in $\RR^D$, while noisy observations are generated by isotropic Gaussian perturbations in the ambient space. Within this data setting, we analyze three geometric scenarios for the support of the clean data: a single smooth manifold, a pairwise separated finite union of smooth manifolds, and a finite union of intersecting smooth manifolds. Our analysis combines manifold projection, local chart representations, partitions of unity, and neural network approximation to characterize how the reconstruction error depends on the training sample size and the geometry of the clean-data support.

We summarize our main results here. For the single manifold scenario, we establish an upper bound on the expected squared reconstruction error consisting of three terms: a sample-dependent statistical term, a residual term due to tangential noise, and an exponentially small tail term associated with failure of the manifold projection. The statistical term scales as $n^{-2/(d+2)}\log^4 n$ up to its dependence on the ambient dimension, so the exponent in the sample size scaling law is governed by the intrinsic dimension $d$ rather than the ambient dimension $D$. The residual term caused by tangential noise is of order $d\sigma^2/D$, which is negligible when $D \gg d$. This shows a key benefit of denoising autoencoders: for high-dimensional data with low intrinsic dimension, tangential noise has little effect on the reconstruction. For a pairwise separated finite union of manifolds, we next show that the same form of the error bound continues to hold, provided that the component manifolds are sufficiently separated so that noisy observations remain associated with the correct component. For a finite union of intersecting manifolds, we further decompose the clean-data support into an intersection neighborhood and a non-intersection region. The non-intersection region retains the same componentwise error behavior, while the intersection neighborhood contributes an additional ambiguity term determined by the probability mass of clean samples near the intersections.

We present numerical experiments on synthetic manifolds and real image datasets in Section~\ref{sec:numerical_results}. On synthetic manifolds with known intrinsic dimension, we examine the predicted sample-size scaling law and study how denoising behavior changes in the presence of manifold intersections, with particular attention to the intersection neighborhood and the non-intersection region. We further conduct experiments on MNIST, Extended Yale B, and OCTMNIST to investigate whether similar low-dimensional sample-size behavior can be observed in real high-dimensional image data, assuming that these datasets satisfy the finite union of intersecting manifolds scenarios considered in our theory.

The rest of the paper is organized as follows. In Section~\ref{sec:preliminaries}, we review preliminaries on manifolds, charts, partitions of unity, and neural networks. Section~\ref{sec:main_results} presents the problem setup and main theoretical results. Section~\ref{sec:numerical_results} contains numerical experiments on simulated and real datasets. Section~\ref{sec:proofs} provides the proofs of the main results, and Appendix~\ref{sec:appendix_lemmas} contains the proofs of auxiliary lemmas.

\textbf{Notation:} We use lower-case letters to denote scalars, lower-case bold letters to denote vectors, upper-case letters to denote matrices and constants, and calligraphic letters to denote manifolds, sets, and function classes. 
Throughout the paper, $\xb$ denotes a clean data point, $\yb$ denotes a noisy observation, and $\bepsilon$ denotes the noise perturbation, typically with $\yb=\xb+\bepsilon$. 

\section{Preliminaries}
\label{sec:preliminaries}

In this section, we review the geometric notions on embedded manifolds and the neural network classes used throughout the paper.

\subsection{Manifolds, charts, and partitions of unity}

We first introduce some definitions and notation about manifolds. More details can be found in \cite{tu2011manifolds,lee2006riemannian}. Let $\cM$ be a $d$-dimensional smooth Riemannian manifold isometrically embedded in $\RR^D$. A chart of $\cM$ defines a local neighborhood and local coordinates on $\cM$.

\begin{definition}
A \emph{chart} of $\cM$ is a pair $(U,\phi)$, where $U\subset \cM$ is an open set and $\phi:U\to\RR^d$ is a homeomorphism onto an open subset $\phi(U)\subset\RR^d$. A \emph{$C^s$ atlas} of $\cM$ is a collection of charts $\{(U_k,\phi_k)\}_{k\in \mathcal K}$ such that $\cM=\bigcup_{k\in \mathcal K}U_k$, and for any $k_1,k_2\in \mathcal K$ with $U_{k_1}\cap U_{k_2}\neq\varnothing$, the transition maps
$$
\phi_{k_1}\circ\phi_{k_2}^{-1}:\phi_{k_2}(U_{k_1}\cap U_{k_2})\to \phi_{k_1}(U_{k_1}\cap U_{k_2}), \qquad
\phi_{k_2}\circ\phi_{k_1}^{-1}:\phi_{k_1}(U_{k_1}\cap U_{k_2})\to \phi_{k_2}(U_{k_1}\cap U_{k_2})
$$
are both of class $C^s$ for all $k_1,k_2\in \mathcal K$. An atlas is \emph{finite} if it contains finitely many charts.
\end{definition}

A smooth manifold is a manifold equipped with a $C^\infty$ atlas. Functions of class $C^s$ on a smooth manifold are defined through their local coordinate representations.

\begin{definition}
Let $\cM$ be a smooth manifold and let $f:\cM\to\RR$ be a function on $\cM$. We say that $f$ is of class $C^s$ on $\cM$ if for every chart $(U,\phi)$ of $\cM$, the function
$f\circ \phi^{-1}:\phi(U)\to\RR$ is of class $C^s$ in the usual Euclidean sense.
\end{definition}

Local charts alone are not enough to build a global construction on $\cM$. To combine local objects defined on different charts, we use partitions of unity.

\begin{definition}
A $C^\infty$ \emph{partition of unity} on $\cM$  is a collection of $C^\infty$ functions $\{\rho_k\}_{k\in \mathcal K}$, where $\rho_k:\cM\to[0,1]$, such that for any $\xb\in\cM$, $\sum_{k\in\mathcal K}\rho_k(\xb)=1$ and every $\xb\in\cM$ has a neighborhood on which only finitely many $\rho_k$ are nonzero.
\end{definition}

An open cover $\{U_k\}_{k\in \mathcal K}$ of $\cM$ is \emph{locally finite} if every $\xb\in\cM$ has a neighborhood that intersects only finitely many sets in the cover. A partition of unity $\{\rho_k\}_{k\in \mathcal K}$ is said to be \emph{subordinate} to $\{U_k\}_{k\in \mathcal K}$ if for every $k\in \mathcal K$, $\operatorname{supp}(\rho_k)\subset U_k$. For any locally finite open cover of a smooth manifold, there exists a $C^\infty$ partition of unity subordinate to that cover \cite[Chapter 2, Theorem 15]{spivak1975comprehensive}.

We next introduce the reach and tubular neighborhood of an embedded manifold, which characterize the region where the orthogonal projection onto the manifold is uniquely defined.

\begin{definition}[Reach \cite{federer1959curvature,niyogi2008finding}]
For any $\yb\in\RR^D$, define $\operatorname{dist}(\yb,\cM) := \inf_{\xb\in\cM}\|\yb-\xb\|_2$. The \emph{medial axis} of $\cM$ is the set
$$
\operatorname{Med}(\cM)
=
\left\{
\yb\in\RR^D :
\exists\,\xb_1\neq\xb_2\in\cM
\text{ such that }
\norm{\yb-\xb_1}_2
=
\norm{\yb-\xb_2}_2
=
\operatorname{dist}(\yb,\cM)
\right\}.
$$
For any $\xb\in\cM$, the \emph{local reach} of $\cM$ at $\xb$ is defined by $\tau_{\cM}(\xb)=\operatorname{dist}\bigl(\xb,\operatorname{Med}(\cM)\bigr)$. The \emph{reach} of $\cM$ is $\tau_{\cM}=\inf_{\xb\in\cM}\tau_{\cM}(\xb)$. For any $q\in[0,1)$, the \emph{relative tubular neighborhood} of radius $q$ is defined by
\begin{equation}
\cM(q)
=
\left\{
\yb=\xb+\ub:
\xb\in\cM,\;
\ub\in T_{\xb}^{\perp}\cM,\;
\norm{\ub}_2<q\,\tau_{\cM}(\xb)
\right\}.
\label{eq:tubular}
\end{equation}
For every $\yb\in\cM(q)$, the nearest-point projection onto $\cM$ is uniquely defined as $\pi(\yb) := \argmin_{\vb\in\cM}\|\yb-\vb\|_2$.
\end{definition}
Geometrically, the reach measures the size of the neighborhood around $\cM$ in which the nearest-point projection remains unique. A larger reach therefore corresponds to a wider projection-valid neighborhood. For example, a hyperplane has infinite reach, while a sphere of radius $r$ has reach $r$. For $\yb\in\cM(q)$, we have $\yb=\pi(\yb)+\wb$ and $\wb\in T_{\pi(\yb)}^\perp\cM$. The existence and uniqueness of this projection follow from \cite{niyogi2008finding,cloninger2021deep}.

\subsection{Neural networks}

We next introduce the neural network classes used throughout the paper. We work with feedforward neural networks (FNNs) with the rectified linear unit activation $\operatorname{ReLU}(a) = \max\{a,0\}$.

\begin{definition}
An $L$-layer FNN with input dimension $d_1$ and output dimension $d_2$ is a mapping $f:\RR^{d_1}\to\RR^{d_2}$ of the form
$$
f(\zb)
=
W_L\,\mathrm{ReLU}\!\Bigl(
W_{L-1}\,\mathrm{ReLU}\bigl(\cdots \mathrm{ReLU}(W_1\zb+\bb_1)+\bb_{L-1}\bigr)
\Bigr)
+\bb_L,
$$
where $W_i$ and $\bb_i$ denote the weight matrix and bias vector of the $i$th layer, respectively.
\end{definition}
We denote the class of ReLU networks with input dimension $d_1$ and output dimension $d_2$ by
$$
\cF(d_1,d_2;L,p,K,\kappa,R)
=
\left\{
f:\RR^{d_1}\to\RR^{d_2} \;\middle|\;
\begin{array}{l}
\text{$f$ has $L$ layers and width at most $p$,} \\[0.3em]
\norm{f}_{L^\infty} \le R, \\[0.3em]
\sum_{i=1}^L \bigl(\norm{W_i}_0+\norm{\bb_i}_0\bigr)\le K, \\ [0.3em]
\norm{W_i}_{\max}\le \kappa,\;
\norm{\bb_i}_{\infty}\le \kappa
\end{array}
\right\}.
$$
Here, $\norm{\cdot}_0$ denotes the number of nonzero entries. For a vector $\bb$, we define $\norm{\bb}_{\infty}:=\max_j |b_j|$, and for a matrix $A$, we define $\norm{A}_{\max}:=\max_{j,k}|A_{jk}|$. For a vector-valued function $f$ defined on a domain $\Omega$, $\norm{f}_{L^\infty(\Omega)} := \sup_{\zb\in\Omega}\norm{f(\zb)}_\infty$. When the domain is clear from context, we simply write
$\norm{f}_{L^\infty}$. This network class will be used to specify the encoder and decoder architectures in the main results.

\section{Denoising autoencoders and generalization error analysis}
\label{sec:main_results}

In this section, we introduce the denoising autoencoder framework for recovering clean data from noisy observations and present our main theoretical results on its generalization error.

\subsection{Denoising autoencoder}

\begin{figure}[t!]
    \centering
    \includegraphics[width=0.9\linewidth]{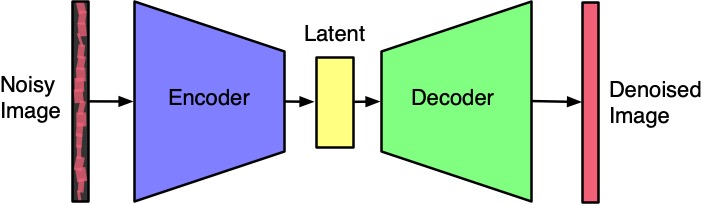}
    \caption{Illustration of a denoising autoencoder. A noisy observation $\yb$ is mapped by the encoder $\cE$ to a latent representation, and the decoder $\cD$ reconstructs a denoised output of the corresponding clean sample $\xb$.}
    \label{fig:dae_overview}
\end{figure}

A denoising autoencoder \cite{vincent2008extracting,bengio2009learning} is trained to reconstruct a clean signal from its corrupted observation. Given a clean sample $\xb$ and a noisy sample $\yb=\xb+\bepsilon$, the encoder maps $\yb$ to a latent representation, and the decoder maps this latent representation back to the data space. Figure~\ref{fig:dae_overview} illustrates the basic architecture of a denoising autoencoder.

The denoising autoencoder is trained by minimizing the reconstruction loss
\begin{equation}
\label{eq:dae_training}
\textstyle\min_{\cE,\cD}
\frac{1}{n}\sum_{i=1}^n
\left\|
\cD\circ\cE(\yb_i)-\xb_i
\right\|_2^2.
\end{equation}
where $\cE$ and $\cD$ are the encoder and decoder networks respectively. Unlike a standard autoencoder, which reconstructs its own input, a denoising autoencoder uses a corrupted input and a clean target. Thus the learned reconstruction map is encouraged to remove the noise component while preserving the underlying data structure. 

\begin{figure}[t!]
    \centering
    \includegraphics[width=0.9\linewidth]{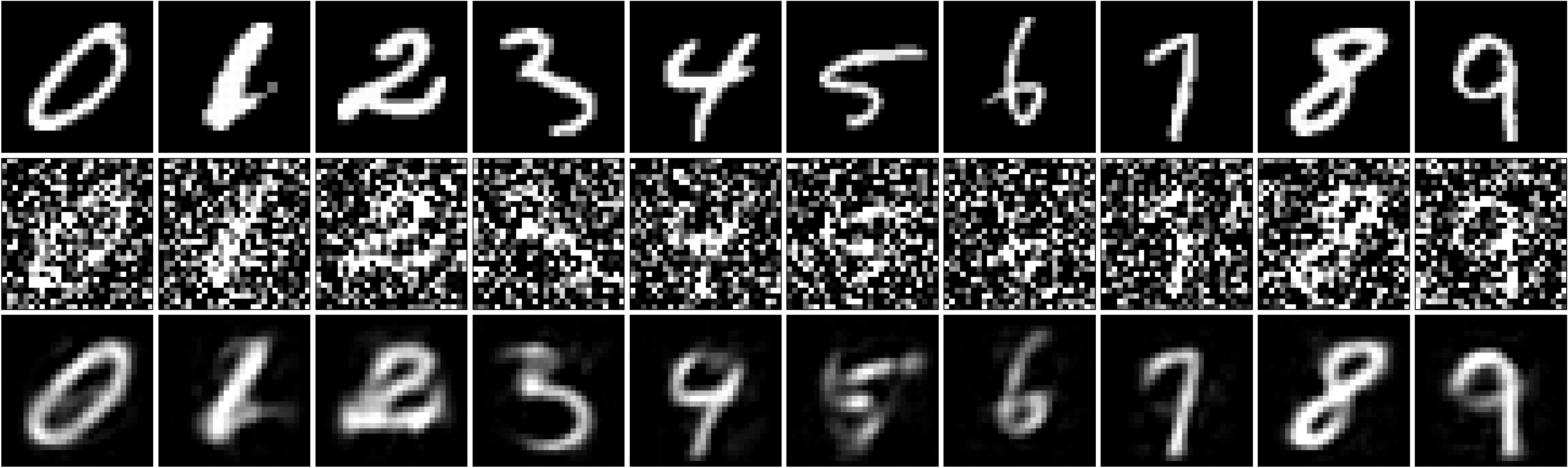}
    \caption{Representative high-noise MNIST denoising results. Columns correspond to digits $0$--$9$, and the rows show clean images (top), noisy observations (middle), and denoised reconstructions (bottom). A single autoencoder is
    trained jointly using all $60{,}000$ MNIST training images. Both training and testing use additive Gaussian noise
    $\varepsilon\sim\mathcal{N}(0,(20^2/784)I_{784})$. The encoder and decoder have architectures $784$--$256$--$128$--$64$ and $64$--$128$--$256$--$784$, respectively. Pixel intensities are displayed
    over the common range $[0,1]$.}
    \label{fig:mnist_dae_demo}
\end{figure}

Figure~\ref{fig:mnist_dae_demo} provides a representative example of denoising on MNIST. The same DAE is trained on noisy and clean image pairs and then applied to previously unseen noisy test images.

\subsection{Data setting}

We now introduce the common data setting used throughout our theoretical analysis. 

\begin{setting}[Data setting]
\label{setting:data}
Let $\mu$ be a probability distribution in $\RR^D$. We consider a training dataset $\cS=\{(\yb_i,\xb_i)\}_{i=1}^n$, where $\xb_i\overset{\mathrm{i.i.d.}}{\sim}\mu$, and the noisy observations are generated by $\yb=\xb+\bepsilon$, where $\bepsilon\sim\mathcal N\!\left(0,\frac{\sigma^2}{D}I_D\right)$. The noise vectors are independent across samples and independent of the corresponding clean samples. The normalization by $D$ ensures that $\EE\|\bepsilon\|_2^2=\sigma^2$, so that $\sigma$ controls the total noise magnitude independently of the ambient dimension. We denote the joint distribution of $(\yb,\xb)$ by $\gamma$.
\end{setting}

\begin{figure}[t!]
    \centering
    \begin{minipage}[t]{0.32\linewidth}
        \centering
        \includegraphics[width=\linewidth]{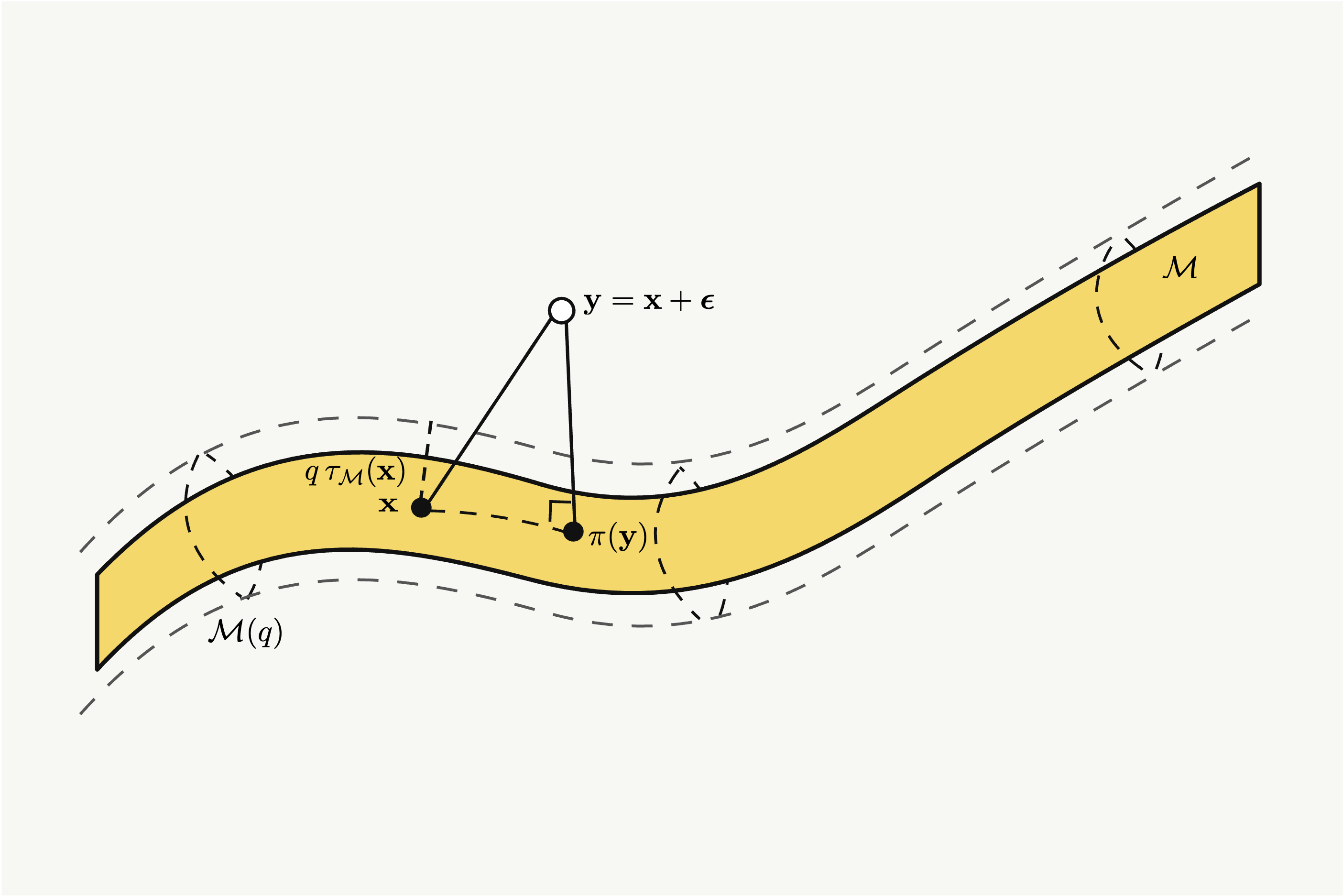}
        \centerline{(a) Single manifold}
    \end{minipage}
    \hfill
    \begin{minipage}[t]{0.32\linewidth}
        \centering
        \includegraphics[width=\linewidth]{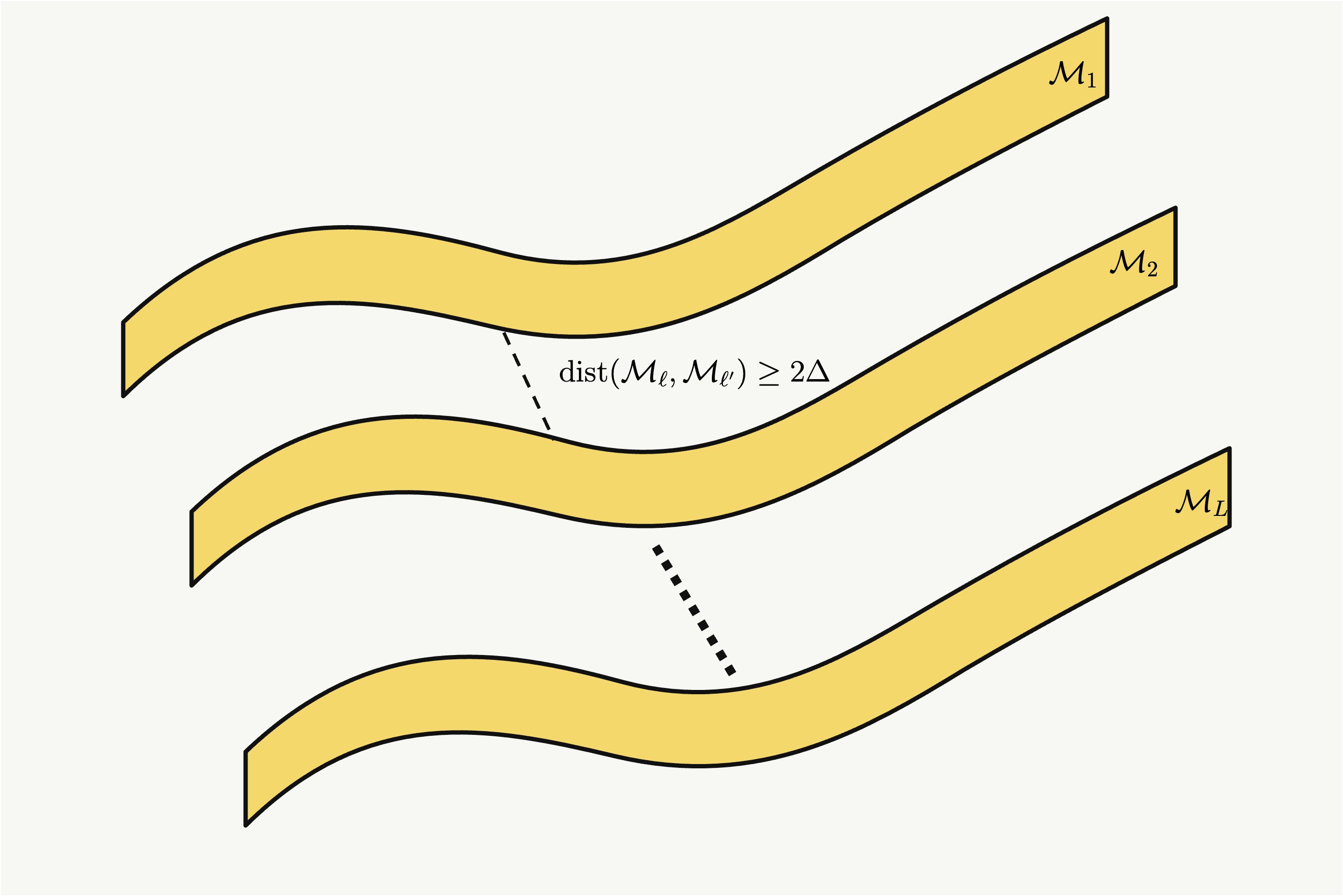}
        \centerline{(b) Pairwise separated union}
    \end{minipage}
    \hfill
    \begin{minipage}[t]{0.32\linewidth}
        \centering
        \includegraphics[width=\linewidth]{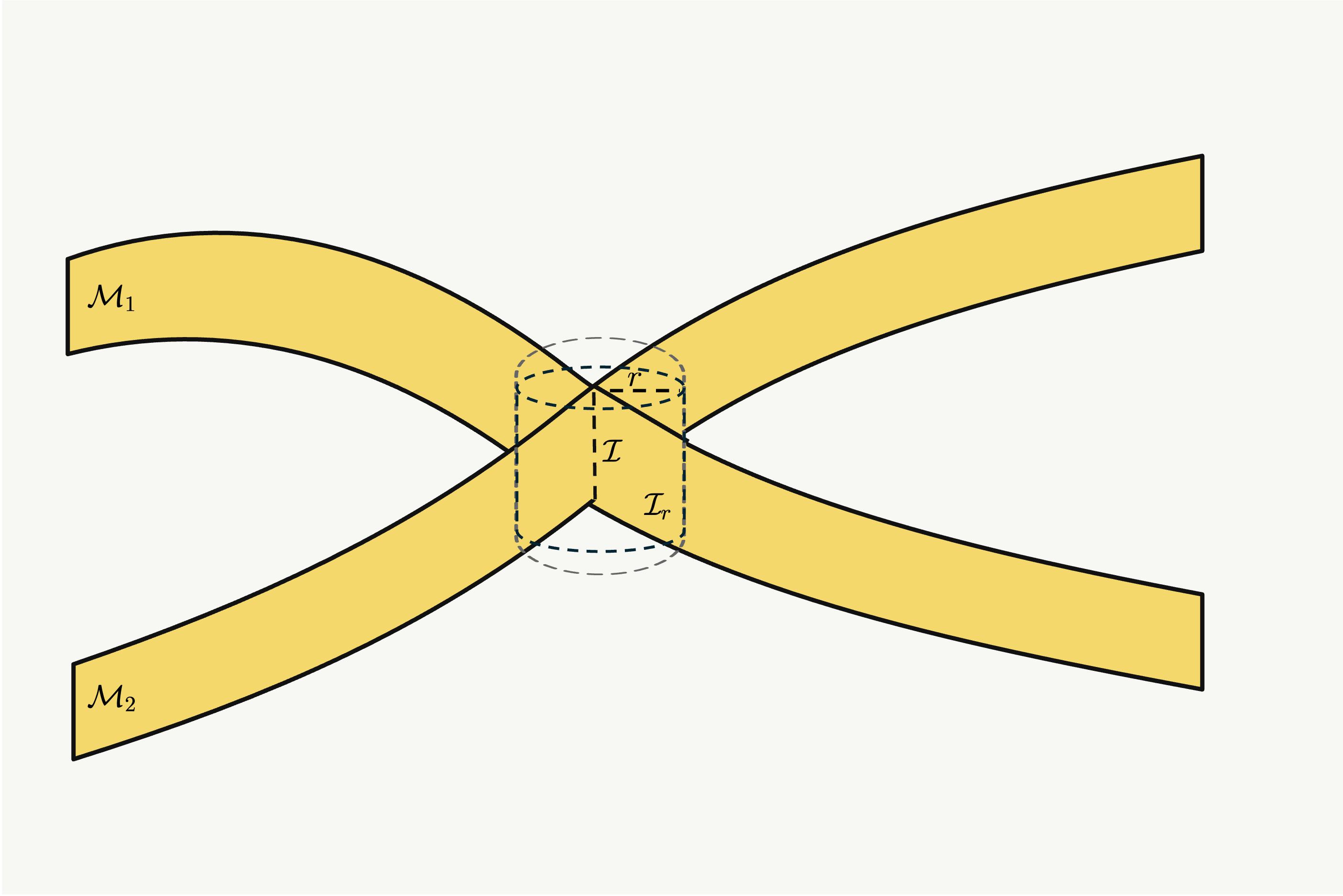}
        \centerline{(c) Intersecting union}
    \end{minipage}
    \caption{Three geometric scenarios considered in the theory. (a) Single manifold: clean samples lie on a smooth manifold $\cM$, and noisy observations are denoised through the local projection. (b) Pairwise separated finite union: the support is $\cM=\bigcup_{\ell=1}^L\cM_\ell$, with pairwise separation ensuring stable component identification under sufficiently small perturbations. (c) Intersecting finite union: the components may intersect on $\cI$, so component identification is stable away from the intersection neighborhood $\cI_r$ but may be ambiguous near $\cI_r$.}
    \label{fig:three_scenarios}
\end{figure}

Given encoder and decoder network classes
$\cF_{\mathrm{NN}}^{\cE}$ and $\cF_{\mathrm{NN}}^{\cD}$,
let $(\widehat{\cD},\widehat{\cE})$ denote a global minimizer of the empirical reconstruction loss in~\eqref{eq:dae_training}, and write $\widehat{\cG}:=\widehat{\cD}\circ\widehat{\cE}$.
We measure its performance by the expected squared reconstruction error
\begin{equation}
\label{eq:reconstruction_error}
\textstyle \EE_{\cS}\EE_{(\yb,\xb)\sim\gamma}
\|\widehat{\cG}(\yb)-\xb\|_2^2.
\end{equation}
Under Setting~\ref{setting:data}, we study three scenarios for the geometric structure of the support of the clean-data distribution $\mu$, as illustrated in Figure~\ref{fig:three_scenarios}. We first consider the single-manifold case in Section~\ref{sec:single_manifold}, then extend the analysis to a pairwise separated finite union of manifolds in Section~\ref{sec:union_manifold}, and finally study a finite union of intersecting manifolds in Section~\ref{sec:intersecting_union_manifold}. These three cases correspond to Figure~\ref{fig:three_scenarios}(a), (b), and (c), respectively.

\subsection{Single manifold}
\label{sec:single_manifold}

We first consider the scenario in which the clean data are supported on a single low-dimensional manifold, as illustrated in Figure~\ref{fig:three_scenarios}(a).

\begin{scenario}[Single manifold]
\label{scenario:single}
Under Setting~\ref{setting:data}, assume that the clean-data distribution $\mu$ is supported on a $d$-dimensional compact smooth Riemannian manifold $\cM$ isometrically embedded in $\RR^D$.
\end{scenario}

This scenario has two basic geometric consequences that are used throughout the analysis:
\begin{itemize}
    \item[(i)] Since $\cM$ is compact, it is bounded. In particular, there exists a constant $B>0$ such that for every $\xb\in\cM$, $\norm{\xb}_\infty \le B$.
    
    \item[(ii)] Since $\cM$ is a compact smooth embedded manifold, it has positive reach \cite[Proposition 14]{thale200850}, denoted by $\tau_{\cM}>0$.
\end{itemize}
A general compact smooth manifold may not admit a global Euclidean parametrization. We therefore represent $\cM$ by finitely many local charts and combine the local representations through a partition of unity. The following lemma gives such a multi-chart construction and quantifies the number of charts required to cover $\cM$.

\begin{lemma}[Multi-chart construction, adapted from {\cite[Lemmas~10 and~12]{liu2024chart}}]\label{lem:multichart_construction}
Under Scenario~\ref{scenario:single}, there exists a finite atlas $\{(U_j,\phi_j)\}_{j=1}^{C_{\cM}}$ and a partition of unity subordinate to this atlas. Moreover, for any $0<r<\tau_{\cM}/2$, the number of charts satisfies
$
C_{\cM}
\le
\frac{|\cM|}
{\cos^d\!\left(\arcsin\frac{r}{2\tau_{\cM}}\right)
|B_r^d|},
$
where $|\cM|$ denotes the volume of $\cM$ and $|B_r^d|$ denotes
the volume of the $d$-dimensional Euclidean ball of radius $r$.
\end{lemma}

\begin{remark}
The quantity $C_{\cM}$ gives an upper bound on the number of local charts needed to represent $\cM$. In particular, taking the covering radius proportional to the reach, as in the construction of \cite{liu2024chart}, shows that the required number of charts is finite and increases as the geometric scale $\tau_{\cM}$ decreases. For manifolds with additional global structure, fewer charts may suffice. For example, if $\cM$ admits a single global parametrization, the construction effectively reduces to a single-chart representation.
\end{remark}

We now state the main generalization error bound for the single-manifold scenario. The multi-chart construction in Lemma~\ref{lem:multichart_construction} determines the latent dimension of the encoder and decoder architectures.

\begin{theorem}\label{thm:single_manifold}
Under Setting~\ref{setting:data} and Scenario~\ref{scenario:single}, fix $q\in(0,1)$ and assume $\sigma<q\tau_{\cM}$. Consider the multi-chart construction in Lemma~\ref{lem:multichart_construction}, and let $(\widehat{\cE},\widehat{\cD})$ be a global minimizer of \eqref{eq:dae_training}. Let $\cF_{\mathrm{NN}}^{\cE}=\cF(D,C_{\cM}(d+1);L_{\cE},p_{\cE},K_{\cE},\kappa_{\cE},R_{\cE})$, where $L_{\cE}=O(\log^2 n+\log D)$, $p_{\cE}=O(Dn^{d/(d+2)})$, $K_{\cE}=O((D\log D)n^{d/(d+2)}\log^2 n)$, $\kappa_{\cE}=O(n^{2/(d+2)})$, and $R_{\cE}=\max\{\tau_{\cM}/4,1\}$. Let $\cF_{\mathrm{NN}}^{\cD}=\cF(C_{\cM}(d+1),D;L_{\cD},p_{\cD},K_{\cD},\kappa_{\cD},R_{\cD})$, where $L_{\cD}=O(\log^2 n+\log D)$, $p_{\cD}=O(Dn^{d/(d+2)})$, $K_{\cD}=O(Dn^{d/(d+2)}\log^2 n+D\log D)$, $\kappa_{\cD}=O(n^{1/(d+2)})$, and $R_{\cD}=B$.
Then we have the following upper bound on the expected squared reconstruction error \eqref{eq:reconstruction_error}:
\begin{equation}
\begin{aligned}
\EE_{\cS}\EE_{(\yb,\xb)\sim\gamma}
\bigl\|
\widehat{\cD}\circ\widehat{\cE}(\yb)-\xb
\bigr\|_2^2
\le\;&
C(D^2\log^3 D)\,n^{-\frac{2}{d+2}}\log^4 n +
C_1\frac{d\sigma^2}{D} \\
&+
C_2
\exp\!\left[
-\frac D4\left(
\sqrt{\frac{2q^2\tau_{\cM}^2}{\sigma^2}-1}-1
\right)^2
\right].
\end{aligned}
\label{eq:errorsingle}
\end{equation}
where
$
C_1=4(1-q)^{-2},
$
$C_2=12DB^2$,
and $C>0$ is independent of $n$ and $D$, and depends only on $d,B,\tau_{\cM},q,C_{\cM}$, and the volume of $\cM$. The constants hidden in the $O(\cdot)$ notation depend on $d,\tau_{\cM},q,B,C_{\cM}$, and the volume of $\cM$.
\end{theorem}

Theorem \ref{thm:single_manifold} has three main implications:

\begin{itemize}
    \item[(i)] \textbf{Statistical term.} The first term in \eqref{eq:errorsingle} is the sample-size-dependent statistical error coming from learning the projection-based denoising map from finitely many noisy-clean training pairs. Up to logarithmic factors, it scales as $n^{-2/(d+2)}$, so the sample-size exponent is governed by the intrinsic dimension $d$ rather than the ambient dimension $D$.

    \item[(ii)] \textbf{Tangential-noise term.}
    The second term in \eqref{eq:errorsingle} is proportional to the tangential noise, which cannot be fully removed by projection-based reconstruction. While the normal component of the ambient noise can be corrected through the geometry of the manifold, the tangential component leads to a residual error of order $d\sigma^2/D$. When $D \gg d$, this term is negligible, so tangential noise has little effect on denoising high-dimensional data with low intrinsic dimension.

    \item[(iii)] \textbf{Projection-failure tail term.}
    The final term accounts for the event that a noisy observation leaves the projection-valid neighborhood of $\cM$, so that the nearest-point projection may fail to be well defined. This term decays exponentially in $D$ when the noise level is below the projection-valid scale.
\end{itemize}

\begin{remark}
The parameter $q\in(0,1)$ specifies the relative projection-valid scale and must satisfy $\sigma<q\tau_{\cM}$. Increasing $q$ reduces the projection-failure tail but enlarges the factor $C_1=4(1-q)^{-2}$ in the tangential-noise term. For fixed $q$ and $\sigma$, when $d\ll D$, the dominant sample-size dependence remains governed by the intrinsic dimension $d$.
\end{remark}

We defer the detailed proof to Section~\ref{proof:single_manifold} and summarize the main five steps below.

\noindent
\textbf{Step 1: Projection-valid event.}
We define the projection-valid event
\begin{equation}
\label{eq:projection_valid_event}
\textstyle
E_q := \left\{\|\bepsilon\|_2<q\tau_{\cM}\right\}.
\end{equation}
On $E_q$, we have $\operatorname{dist}(\yb,\cM)\le\|\bepsilon\|_2<q\tau_{\cM}$. Hence the nearest-point projection $\pi(\yb)$ is uniquely defined and $\yb\in\cM(q)$. The probability of $E_q^c$ is controlled by a Gaussian norm tail bound.

\noindent
\textbf{Step 2: Error decomposition.}
Following the generalization-error decomposition in \cite{liu2024chart}, we decompose the expected squared reconstruction error \eqref{eq:reconstruction_error} into a bias term and a variance term. The bias term captures the approximation error of the chart-based reconstruction map, while the variance term captures the stochastic error arising from empirical minimization.

\noindent
\textbf{Step 3: Bounding the bias term.}
On the projection-valid event $E_q$, we construct a multi-chart oracle encoder and decoder pair whose composition recovers the manifold projection $\pi(\yb)$. This construction follows the projection-based framework of \cite{cloninger2021deep} and the multi-chart partition-of-unity construction of \cite{liu2024chart}. We then approximate the oracle encoder and decoder by neural networks with accuracy $\varepsilon$. The remaining discrepancy between $\pi(\yb)$ and the clean sample $\xb$ is controlled by the tangential component of the noise.

\noindent
\textbf{Step 4: Bounding the variance term.}
We control the variance term using the covering-number argument in \cite[Lemmas~2 and~3]{liu2024chart}. The variance is first bounded by the covering number of the induced reconstruction class, which is then controlled by the encoder and decoder network architectures.

\noindent
\textbf{Step 5: Combining the bounds.}
We finally combine the upper bounds for the bias term, the variance term, and the projection-failure tail contribution. After balancing the approximation accuracy and the network size, we obtain Theorem~\ref{thm:single_manifold}.

\subsection{Finite union of pairwise separated manifolds}
\label{sec:union_manifold}

We next consider the case where the clean-data distribution is supported on a finite union of pairwise separated manifold components, as illustrated in Figure~\ref{fig:three_scenarios}(b). Compared with the single-manifold scenario, the main additional requirement is that the noisy observation remain associated with the same component as its clean counterpart.

\begin{scenario}[Finite union of pairwise separated manifolds]
\label{scenario:separated} Under Setting~\ref{setting:data}, assume that the clean-data distribution $\mu$ is supported on $\cM = \bigcup_{\ell=1}^L \cM_\ell$, where each $\cM_\ell\subset\RR^D$ is a compact smooth $d$-dimensional Riemannian manifold. Let $\tau_* := \min_{1\le \ell\le L}\tau_{\cM_\ell}>0$. Assume that the components are pairwise separated: there exists $\Delta>\tau_*$ such that for every $\ell\neq\ell'$, $\operatorname{dist}(\cM_\ell,\cM_{\ell'})
\ge 2\Delta$.
\end{scenario}
Since $\cM=\bigcup_{\ell=1}^L \cM_\ell$ is compact, there exists $B>0$ such that for every $\xb\in\cM$, $\norm{\xb}_\infty\le B$. The quantity $\tau_*$ provides a uniform lower bound on the reach of the component manifolds, while the separation condition ensures that sufficiently small perturbations cannot move a sample closer to a different component. The finite atlases of the component manifolds can therefore be combined into a finite atlas for the union, with total number of charts $\sum_{\ell=1}^L C_{\cM_\ell}$. This allows the multi-chart construction from Scenario~\ref{scenario:single} to be applied componentwise once stable component identification is established.

\begin{theorem}\label{thm:union_manifold}
Under Setting~\ref{setting:data} and Scenario~\ref{scenario:separated}, fix $q\in(0,1)$ and assume $\sigma<q\tau_*$. Let $(\widehat{\cE},\widehat{\cD})$ be a global minimizer of \eqref{eq:dae_training}. Let $\cF_{\mathrm{NN}}^{\cE}=\cF(D,\sum_{\ell=1}^L C_{\cM_\ell}(d+1);L_{\cE},p_{\cE},K_{\cE},\kappa_{\cE},R_{\cE})$, where $L_{\cE}=O(\log^2 n+\log D)$, $p_{\cE}=O(Dn^{d/(d+2)})$, $K_{\cE}=O((D\log D)n^{d/(d+2)}\log^2 n)$, $\kappa_{\cE}=O(n^{2/(d+2)})$, and $R_{\cE}=\max\{\frac{\tau_*}{4},1\}$. Let $\cF_{\mathrm{NN}}^{\cD}=\cF(\sum_{\ell=1}^L C_{\cM_\ell}(d+1),D;L_{\cD},p_{\cD},K_{\cD},\kappa_{\cD},R_{\cD})$, where $L_{\cD}=O(\log^2 n+\log D)$, $p_{\cD}=O(Dn^{d/(d+2)})$, $K_{\cD}=O(Dn^{d/(d+2)}\log^2 n+D\log D)$, $\kappa_{\cD}=O(n^{1/(d+2)})$, and $R_{\cD}=B$. Here $C_{\cM_\ell}$ is the chart-number bound from Lemma~\ref{lem:multichart_construction} applied to $\cM_\ell$ using $\tau_*$ as a common lower bound on the component reaches. Then we have the following upper bound on the expected squared reconstruction error \eqref{eq:reconstruction_error}:
\begin{equation}
\begin{aligned}
\EE_{\cS}\EE_{(\yb,\xb)\sim\gamma}
\bigl\|
\widehat{\cD}\circ\widehat{\cE}(\yb)-\xb
\bigr\|_2^2
\le\;&
C(D^2\log^3 D)\,n^{-2/(d+2)}\log^4 n \\
&+
C_1\frac{d\sigma^2}{D} +
C_2
\exp\!\left[
-\frac D4\left(
\sqrt{\frac{2q^2\tau_*^2}{\sigma^2}-1}-1
\right)^2
\right].
\end{aligned}
\label{eq:error_union}
\end{equation}
where $C_1=4(1-q)^{-2}$,$C_2=12DB^2$ and $C>0$ is independent of $n$ and $D$, and depends on $d,q,B,\tau_*,\Delta,L,\{C_{\cM_\ell}\}_{\ell=1}^L$, and the geometric parameters of the component manifolds. The constants hidden in the $O(\cdot)$ notation depend on the same quantities.
\end{theorem}

Theorem~\ref{thm:union_manifold} has the same three-term error structure as Theorem~\ref{thm:single_manifold}. The main differences are that the effective reach is $\tau_*$, the latent dimension depends on the total number of charts across all components, and the separation condition ensures stable component identification under noise.

\begin{remark}
If $\cM_\ell$ has intrinsic dimension $d_\ell$ and component probability $p_\ell\ge p_{\min}>0$, the construction applies componentwise with latent dimension $C_{\cM_\ell}(d_\ell+1)$. Let $d_{\max}:=\max_{1\le\ell\le L}d_\ell$. The worst-case statistical rate is then $n^{-2/(d_{\max}+2)}$ up to logarithmic factors, while the tangential-noise term is of order $d_{\max}\sigma^2/D$.
\end{remark}

\begin{remark}[Interpretation of noise level in applications]
In applications, image intensities are typically normalized to a fixed range, such as $[0,1]$ or $[-1,1]$, before Gaussian noise is added. In our experiments, all images are normalized to $[0,1]$. Our results can be applied directly to the normalized data manifold. For example, if $\cM\subset[-B,B]^D$, rescaling $\zb=\xb/B$ gives a manifold $\cM/B\subset[-1,1]^D$. Adding noise $\bepsilon\sim\mathcal N(0,\sigma^2 I_D/D)$ to $\zb$ corresponds to noise with covariance $B^2\sigma^2 I_D/D$ in the original coordinates. Thus, $\sigma$ measures the noise magnitude relative to the chosen intensity scale. Meanwhile, normalization rescales the manifold reach to $\tau_{\cM}/B$, so the projection-valid condition becomes $\sigma<q\tau_{\cM}/B$. Therefore, normalization does not eliminate the geometric restriction on the noise level.
\end{remark}

We defer the detailed proof to Section~\ref{proof:union_manifold}.

\subsection{Intersecting finite union of manifolds}
\label{sec:intersecting_union_manifold}
\begin{figure}[h]
    \centering
    \includegraphics[width=0.85\linewidth]{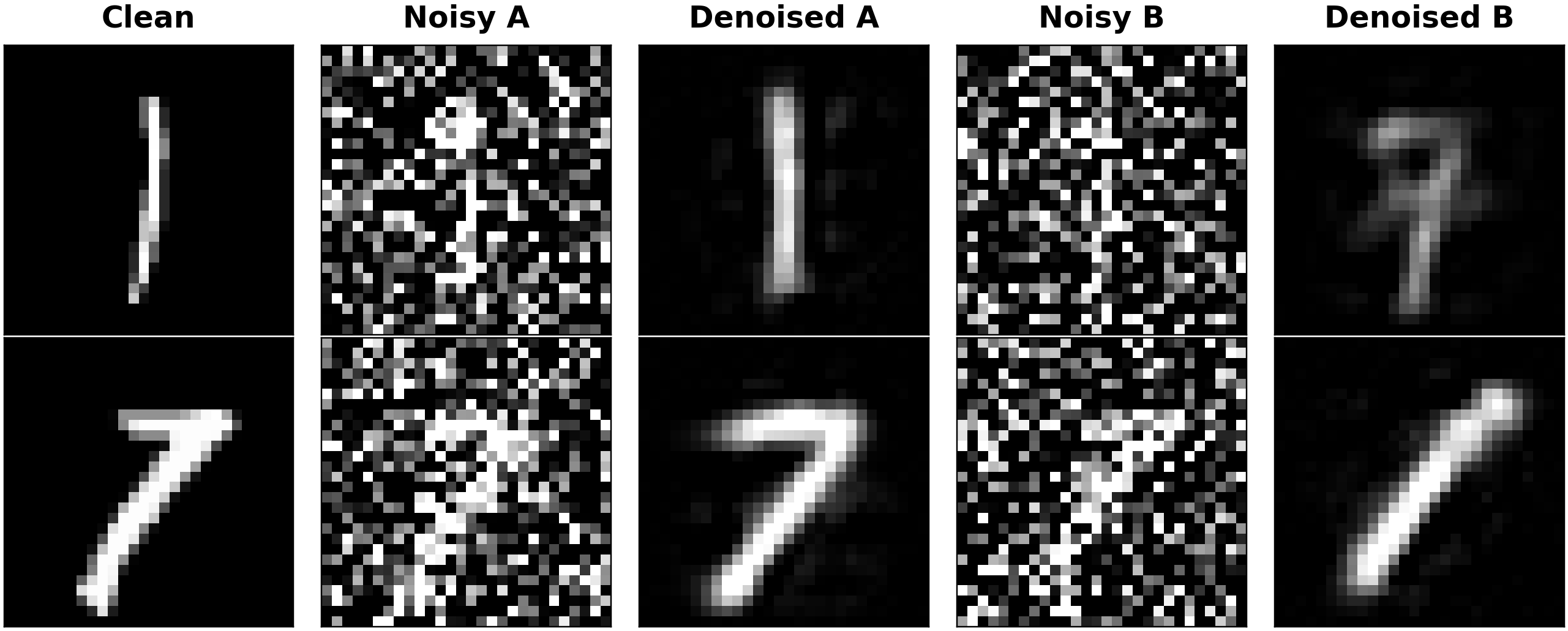}
    \caption{MNIST examples illustrating ambiguity between the class-dependent structures of digits $1$ and $7$. For each clean image, Noisy A and B are independent Gaussian perturbations at the same noise level, $\sigma=20$. Although both are denoised by the same autoencoder, the reconstructions resemble different digit classes.}
    \label{fig:mnist_17_ambiguity}
\end{figure}

The pairwise separation assumption in Scenario~\ref{scenario:separated} may be restrictive for multiclass or multimodal data, where different low-dimensional components can intersect or overlap. Figure~\ref{fig:mnist_17_ambiguity} provides a concrete illustration of the resulting component ambiguity. For the same clean digit, two independent noisy observations at the same noise level can lead to denoised reconstructions associated with different class-dependent structures: a clean digit $1$ may be reconstructed as either $1$ or $7$, and similarly for a clean digit $7$. This suggests that when different low-dimensional structures overlap, the component identity of a clean sample may not be stably recovered from its noisy observation.

Figure~\ref{fig:three_scenarios}(c) illustrates the idea for our analysis. Away from the intersection set, the component manifolds remain locally separated, so the argument from Scenario~\ref{scenario:separated} can still be applied. Near the intersection set, however, different components meet or become close to each other, and the component identity may be ambiguous under noise. This suggests treating the intersection neighborhood and the non-intersection region separately in the error analysis. 

\begin{scenario}[Intersecting finite union]
\label{scenario:intersecting}
Under Setting~\ref{setting:data}, assume the clean-data distribution $\mu$ is supported on $\cM=\bigcup_{\ell=1}^L \cM_\ell$, where each $\cM_\ell\subset\RR^D$ is a compact smooth $d$-dimensional Riemannian manifold. Let $\tau_* = \min_{1\le \ell\le L}\tau_{\cM_\ell}>0$. Assume that at least two component manifolds intersect, and let $\cI = \bigcup_{1\le\ell<\ell'\le L} (\cM_\ell\cap\cM_{\ell'}) \neq\varnothing$ be the intersection set.
\end{scenario}
Define the intersection neighborhood on the clean-data by $\cI_r := \left\{\xb\in\cM: \operatorname{dist}(\xb,\cI)\le r \right\}$ for $r > 0$. Let $p_r:=\PP(\xb\in\cI_r)$ denote the probability mass of the intersection neighborhood, which measures the amount of clean-data mass near the intersections. For each component, define the non-intersection region $
\cM_{\ell,r} := \left\{ \xb\in\cM_\ell: \operatorname{dist}(\xb,\cI)>r \right\}$. We assume that these regions are separated from the other components and define $2\Delta_r := \inf_{\ell\neq\ell'} \operatorname{dist}(\cM_{\ell,r},\cM_{\ell'})$. Assume $\Delta_r>0$. Let $\rho_r := \min\left\{ q\tau_*, \frac{\Delta_r}{2} \right\}$. The radius $\rho_r$ is the effective perturbation scale that guarantees both projection validity and stable component identification in the non-intersection region.

\begin{theorem}[Intersecting union with an ambiguity term]
\label{thm:intersecting_union}
Under Setting~\ref{setting:data} and
Scenario~\ref{scenario:intersecting}, fix $q\in(0,1)$ and $r>0$.
Assume $\Delta_r>0$ and $\rho_r>\sigma$.
Let $\widehat{\cE},\widehat{\cD}$ be a global minimizer in
\eqref{eq:dae_training} with network classes having the same $n$- and $D$-scaling as in Theorem~\ref{thm:union_manifold} Then the expected squared reconstruction error satisfies
\begin{equation}
\label{eq:intersecting_union_bound}
\begin{aligned}
\EE_{\cS}\EE_{(\yb,\xb)\sim\gamma}
\bigl\|
\widehat{\cD}\circ\widehat{\cE}(\yb)-\xb
\bigr\|_2^2
&\le
C(D^2\log^3 D)n^{-2/(d+2)}\log^4 n
+
C_1(1-p_r)\frac{d\sigma^2}{D}
\\
&\quad+
C_2
\exp\!\left[
-\frac D4\left(
\sqrt{\frac{2\rho_r^2}{\sigma^2}-1}-1
\right)^2
\right]
+
8DB^2p_r,
\end{aligned}
\end{equation}
where $C_1=4(1-q)^{-2}$, $C_2=12DB^2$, and $C>0$ is independent of $n$ and $D$.
The constant may depend on
$d,q,B,\tau_*,r,\Delta_r,L,
\{C_{\cM_\ell}\}_{\ell=1}^L$,
and the geometric parameters of the component manifolds.
\end{theorem}

We defer the proof to Section~\ref{proof:intersecting_union}.
Theorem~\ref{thm:intersecting_union} separates the reconstruction
error into the contribution from the non-intersection region and the
additional ambiguity caused by samples near the intersections.

\begin{itemize}
    \item[(i)] \textbf{Non-intersection region.}
    Away from the intersection set, the tangential-noise term retains the same componentwise form as in Theorem~\ref{thm:union_manifold} and is
    weighted by the probability $1-p_r$ of the non-intersection region.
    The projection-failure tail term is controlled globally through the
    Gaussian noise tail, while the statistical term remains a global
    estimation term with rate $n^{-2/(d+2)}$ up to logarithmic factors.

    \item[(ii)] \textbf{Intersection neighborhood.}
    Near the intersection set, the component identity may be
    ambiguous. Since the reconstruction error is bounded on this
    region, its contribution is controlled by
    $8DB^2p_r$, which is proportional to the probability mass of
    clean samples near the intersections.
\end{itemize}

\section{Numerical Results}
\label{sec:numerical_results}

In this section, we present numerical experiments corresponding to the three geometric scenarios considered in our theory. We first study synthetic manifolds with known intrinsic dimension to examine the predicted sample-size scaling law. We then consider intersecting synthetic manifolds to compare the reconstruction behavior in the intersection neighborhood and the non-intersection region. Finally, we study the real image datasets, including MNIST \cite{lecun1998gradient}, Extended Yale B \cite{georghiades2001few}, and OCTMNIST \cite{yang2023medmnist}, to examine whether similar low-dimensional sample-size behavior is observed in high-dimensional image data. 

All experiments in this section use a standard fully connected denoising autoencoder rather than the full multi-chart partition-of-unity architecture used in the theoretical analysis. Thus, the experiments provide a simple baseline for examining the qualitative predictions of the theory. Despite using this simpler architecture, the numerical results already exhibit the intrinsic-dimension-dependent sample-size behavior predicted by our analysis. 

\subsection{Single synthetic manifolds}
\label{sec:single_synthetic}

We first consider single synthetic manifolds with controlled geometry, the circle $S^1$ with $(D,d)=(6,1)$, the sphere $S^2$ with $(D,d)=(6,2)$, the $3$-sphere $S^3$ with $(D,d)=(6,3)$, a genus--2 surface with $(D,d)=(3,2)$, and the swiss roll with $(D,d)=(3,2)$. Their intrinsic dimensions are known, allowing direct comparison between the empirical sample-size scaling and the theoretical rate $n^{-\frac{2}{d+2}}$.

\begin{figure}[t!]
    \centering
    \begin{minipage}[t]{0.48\linewidth}
        \centering
        \includegraphics[width=\linewidth]{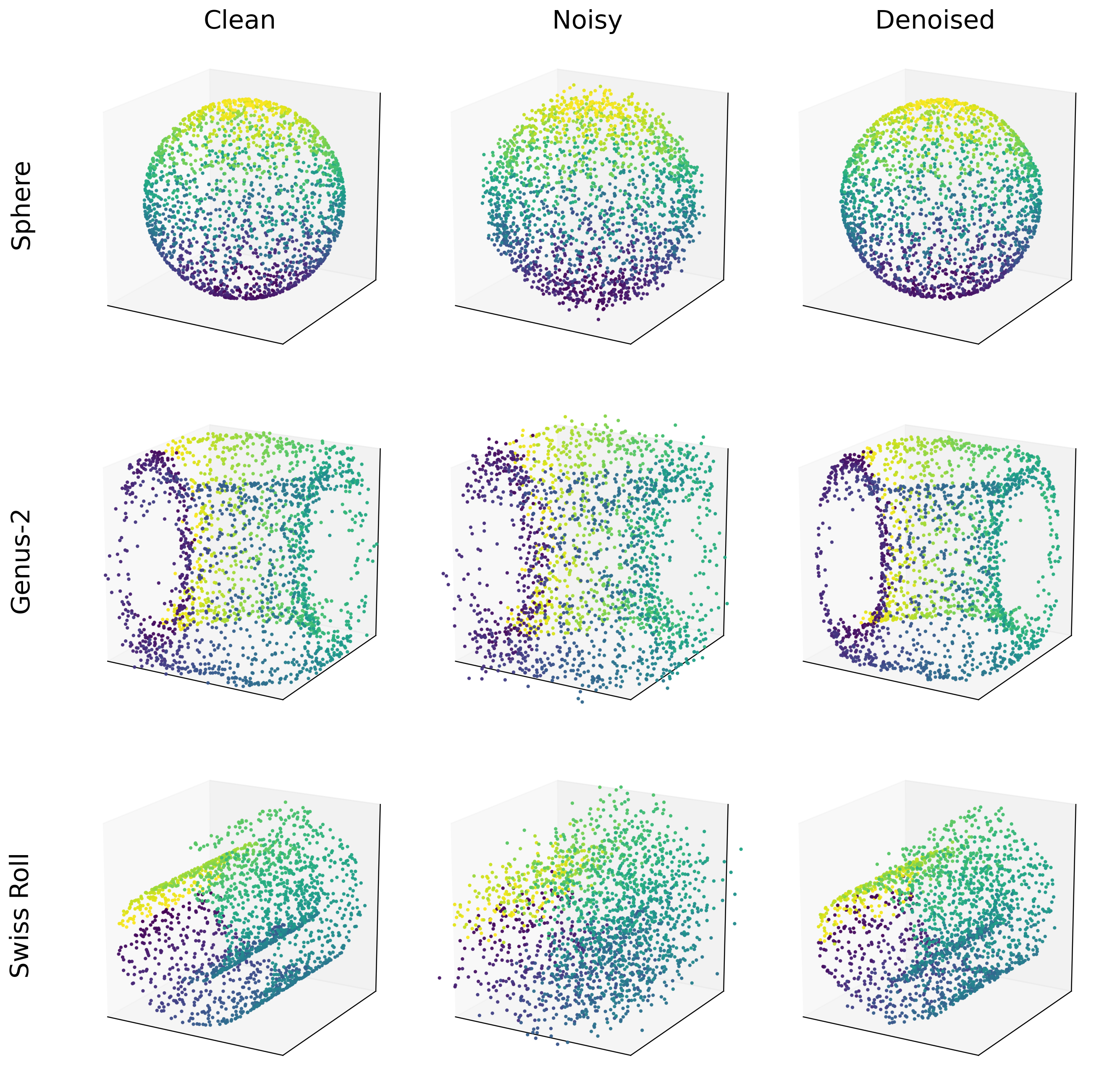}
    \end{minipage}
    \hfill
    \begin{minipage}[t]{0.48\linewidth}
        \centering
        \includegraphics[width=\linewidth]{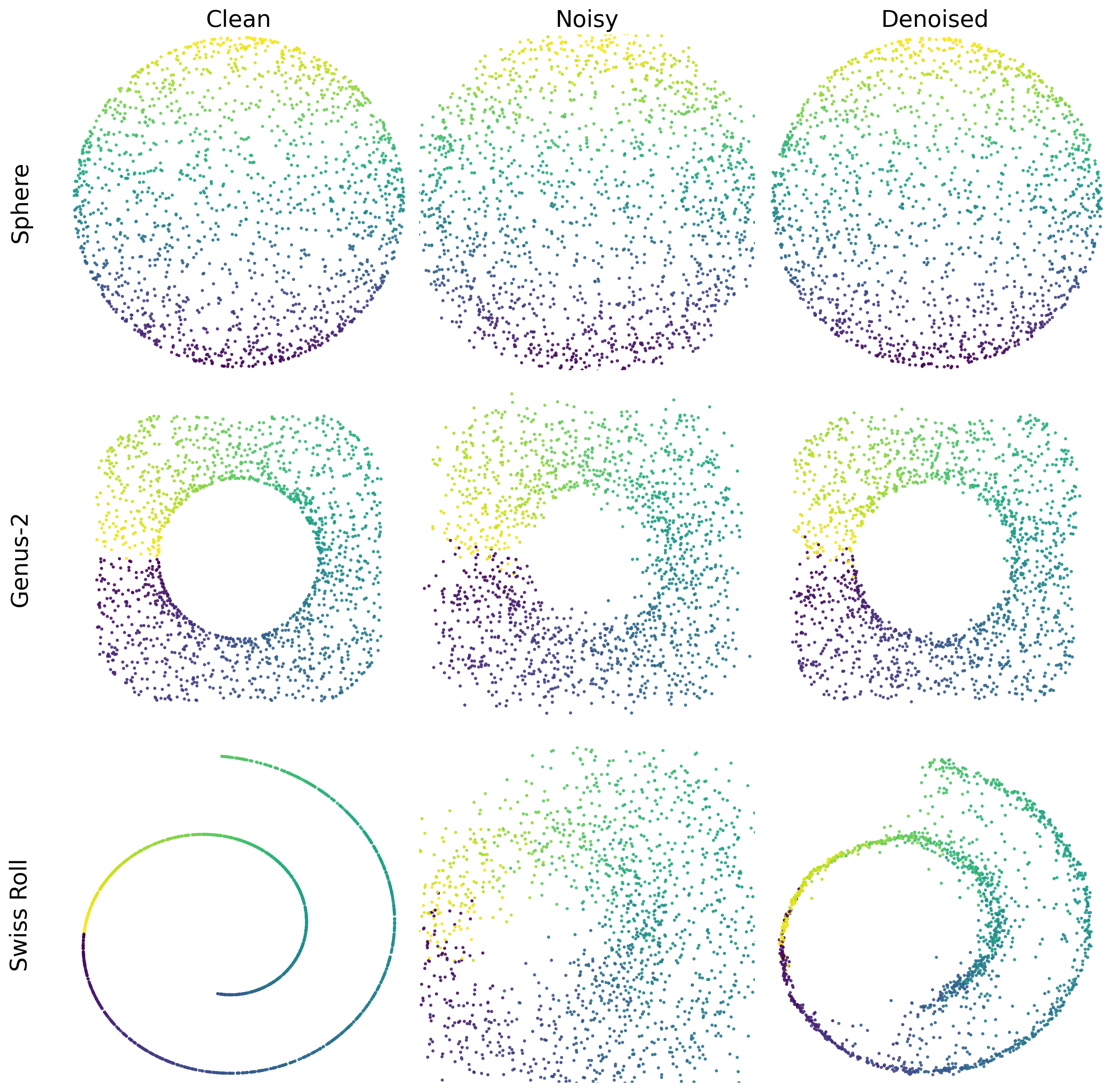}
    \end{minipage}
    \caption{Qualitative denoising results on representative single synthetic manifolds. Rows correspond to the sphere, genus--2 surface, and swiss roll, and columns show clean samples, noisy observations, and denoised reconstructions. The left block shows three-dimensional views, while the right block shows fixed two-dimensional projections for visualization. Colors encode the same intrinsic coordinates across clean, noisy, and denoised samples.}
    \label{fig:simulated_manifolds}
\end{figure}

We first present representative denoising results on several synthetic manifolds. For the examples embedded in ambient dimension $D=6$, clean samples are generated in their natural coordinates and then mapped into $\mathbb{R}^6$ by a fixed orthonormal embedding. Figure~\ref{fig:simulated_manifolds} shows representative test samples from three datasets. In each row, the clean samples are corrupted by Gaussian noise and then processed by the trained denoising autoencoder. The denoised samples move back toward the underlying geometric structure, consistent with the projection-based interpretation in the theory.

\begin{figure}[h]
    \centering
    \includegraphics[width=0.9\linewidth]
        {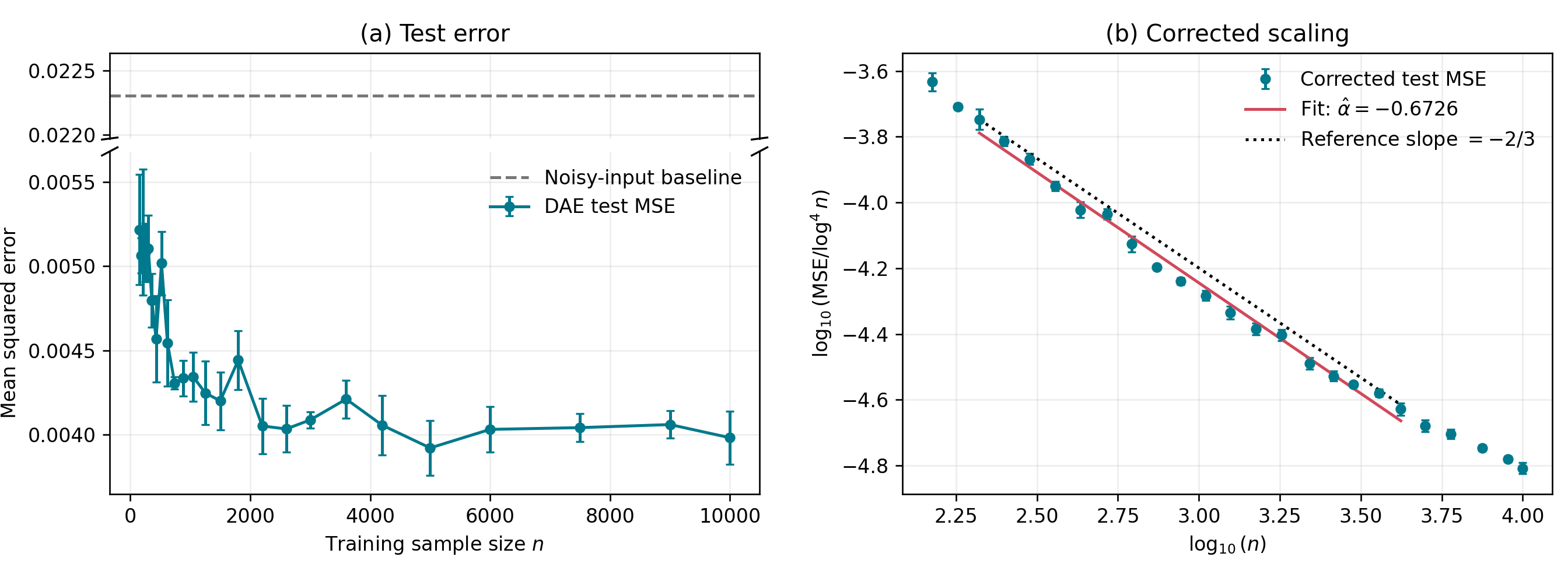}
    \caption{Sample-size scaling on the circle dataset with ambient
    dimension $D=6$ and intrinsic dimension $d=1$.
    Left: test MSE versus training sample size $n$, together with the
    noisy-input baseline.
    Right: corrected log--log plot of
    $\mathrm{MSE}(n)/\log^4 n$ and the fitted slope over the
    pre-floor regime.
    The fitted slope is close to the theoretical value
    $\alpha_{\mathrm{theory}}=-2/(d+2)=-2/3$.}
    \label{fig:circle_scaling}
\end{figure}

Figure~\ref{fig:circle_scaling} illustrates the sample-size scaling procedure on the circle dataset. To estimate the empirical sample-size rate, we restrict the fit to the pre-floor regime. Since the sample-size-dependent statistical term in the theory contains the factor $\log^4 n$, we define the corrected error $\mathrm{MSE}_{\mathrm{corr}}(n) = \frac{\mathrm{MSE}(n)}{\log^4 n}$. 
Ignoring the residual floor, the theory predicts
$
\mathrm{MSE}_{\mathrm{corr}}(n)
\propto
n^{-\frac{2}{d+2}}
$.
We define the theoretical log--log slope by $\alpha_{\mathrm{theory}}
:=
-\frac{2}{d+2}$.
Hence $\log \mathrm{MSE}_{\mathrm{corr}}(n) = C+\alpha_{\mathrm{theory}}\log n$. We estimate the empirical slope
$\hat{\alpha}_{\mathrm{empirical}}$
from a linear fit of
$\log \mathrm{MSE}_{\mathrm{corr}}(n)$
against $\log n$. For the circle dataset, where $d=1$, the theoretical rate is $n^{-2/3}$, corresponding to $\alpha_{\mathrm{theory}}=-2/3$, which is close to the fitted slope.

Each experiment is repeated over five independent random seeds, and the reported MSE values are averaged across runs. Table~\ref{tab:synthetic_scaling} summarizes the theoretical and empirical log--log slopes. The latent dimension is set to $d+1$, and the fitting range is chosen before the empirical error floor becomes dominant. The empirical slopes are close to the theoretical values across the synthetic examples. In particular, manifolds with the same intrinsic dimension exhibit similar exponents despite different global geometries, supporting the prediction that the dominant sample-size scaling is governed primarily by intrinsic rather than ambient dimension.

\begin{table}[!htbp]
\centering
\small
\renewcommand{\arraystretch}{1.15}
\begin{tabular}{lcccccc}
\toprule
Dataset
& $D$
& $d$
& $\sigma$
& $q_{0.99}/\hat{\tau}$
& $\alpha_{\mathrm{theory}}$
& $\hat{\alpha}_{\mathrm{empirical}}$ \\
\midrule
circle ($S^1$)   & 6 & 1 & 0.149 & $1/4$ & -0.67 & -0.67 \\
sphere ($S^2$)   & 6 & 2 & 0.149 & $1/4$ & -0.50 & -0.50 \\
3-sphere ($S^3$) & 6 & 3 & 0.299 & $1/2$ & -0.40 & -0.48 \\
genus--2 surface & 3 & 2 & 0.151 & $3/4$ & -0.50 & -0.53 \\
swiss roll       & 3 & 2 & 0.259 & $3/4$ & -0.50 & -0.53 \\
\bottomrule
\end{tabular}
\caption{Single synthetic manifolds: theoretical and empirical log--log slopes. Here $\hat{\tau}$ denotes the exact reach for the circle and spheres and the estimated reach for the genus--2 surface and swiss roll. The quantity $q_{0.99}$ satisfies $\PP(\|\bepsilon\|_2\le q_{0.99})=0.99$.}
\label{tab:synthetic_scaling}
\end{table}

\subsection{Intersecting synthetic manifolds}
\label{sec:intersecting_synthetic}

We next examine the intersecting-union setting in
Section~\ref{sec:intersecting_union_manifold}. Theorem~\ref{thm:intersecting_union} predicts two qualitatively different regimes: in the non-intersection region, the usual componentwise denoising behavior should remain visible, yet in the intersection neighborhood, component ambiguity may produce an additional reconstruction error.

\begin{wrapfigure}{r}{0.56\textwidth}
    \centering
    \includegraphics[width=\linewidth]{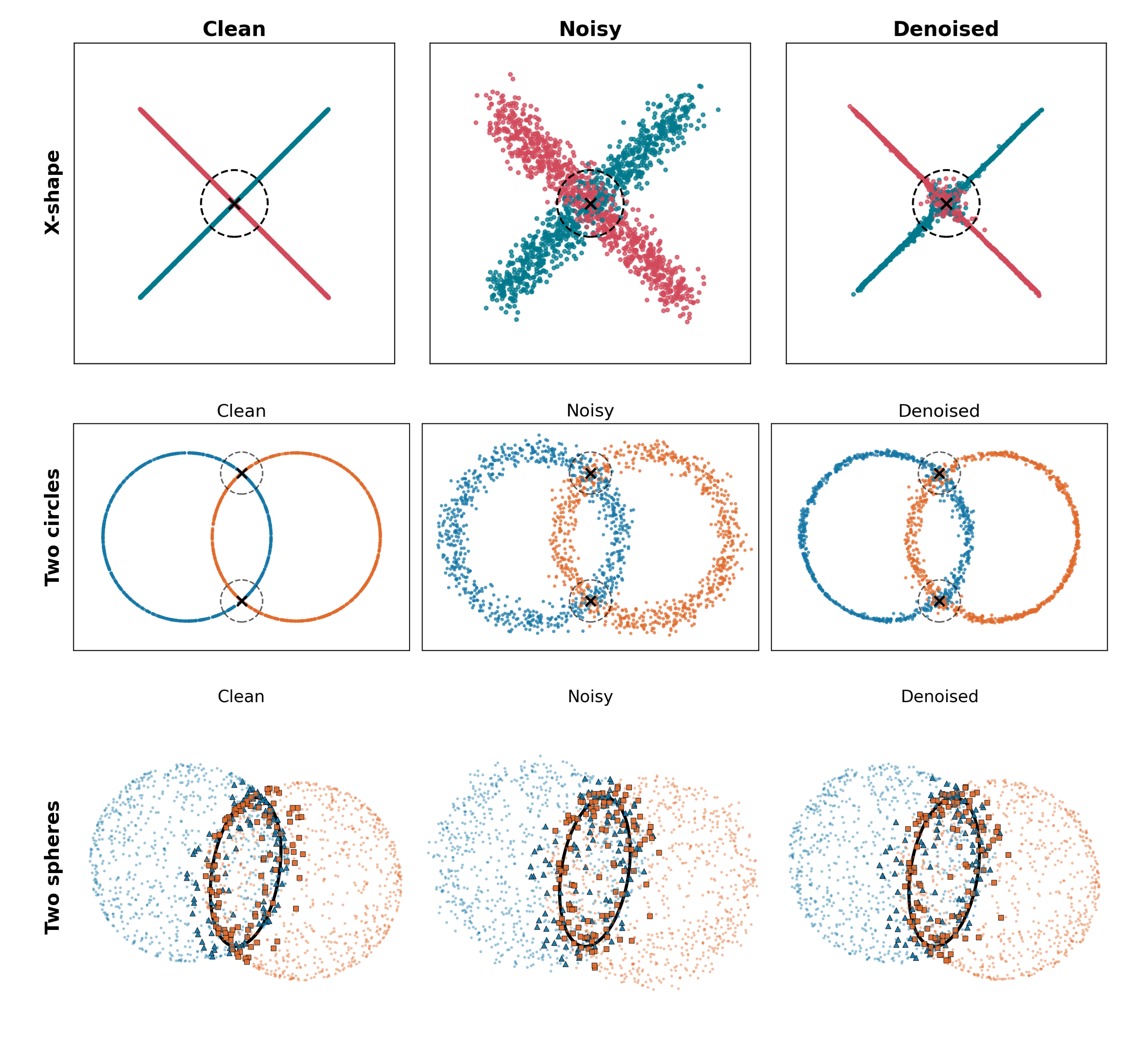}
    \caption{Representative denoising results on intersecting synthetic manifolds. Rows show X-shaped lines, two intersecting circles, and two intersecting $2$-spheres; columns show clean, noisy, and denoised samples. Colors indicate source components, and black markers indicate intersections or their neighborhoods.}
    \label{fig:intersecting_reconstructions}
\end{wrapfigure}

Figure~\ref{fig:intersecting_reconstructions} shows three synthetic examples of increasing geometric complexity: an X-shaped union of two line segments in $\RR^2$, two intersecting unit circles in $\RR^2$, and two intersecting unit $2$-spheres in $\RR^3$. Their component dimensions are $d=1,1,$ and $2$, while the corresponding intersection dimensions are $0,0,$ and $1$, respectively.

For each dataset, we generate noisy-clean training pairs using additive Gaussian noise and train the same fully connected denoising autoencoder as in the preceding experiments, with latent dimension $d+1$. We examine the qualitative reconstruction of the underlying union, the difference between the intersection neighborhood and the non-intersection region, and the sample-size scaling in these two regions.

To quantify the local effect of the intersection, we use the same intersection-neighborhood notation $\cI_r$ introduced in Section~\ref{sec:intersecting_union_manifold}. Samples with $\xb\in\cI_r$ are treated as belonging to the intersection neighborhood, while samples with $\xb\notin\cI_r$ belong to the non-intersection region.  The corresponding regional errors are
$\mathrm{MSE}_{\mathrm{int}} =
\mathbb{E}\!\left[
\|\widehat{\xb}-\xb\|_2^2
\,\middle|\,
\xb\in\mathcal{I}_r
\right]$ and $\mathrm{MSE}_{\mathrm{non}}
=
\mathbb{E}\!\left[
\|\widehat{\xb}-\xb\|_2^2
\,\middle|\,
\xb\notin\mathcal{I}_r
\right]$,
where $\widehat{\xb}=\widehat{\cG}(\yb)$. Because the source component of every synthetic sample is known, we also measure component ambiguity directly. For a sample $\xb$ generated from component $\mathcal{M}_k$, we say that a component switch occurs if the denoised reconstruction $\widehat{\xb}$ is closer to another component than to $\cM_k$. We estimate the component-switch probability separately in the intersection neighborhood and in the non-intersection region.

\begin{figure}[!htbp]
    \centering
    \includegraphics[width=0.9\linewidth]{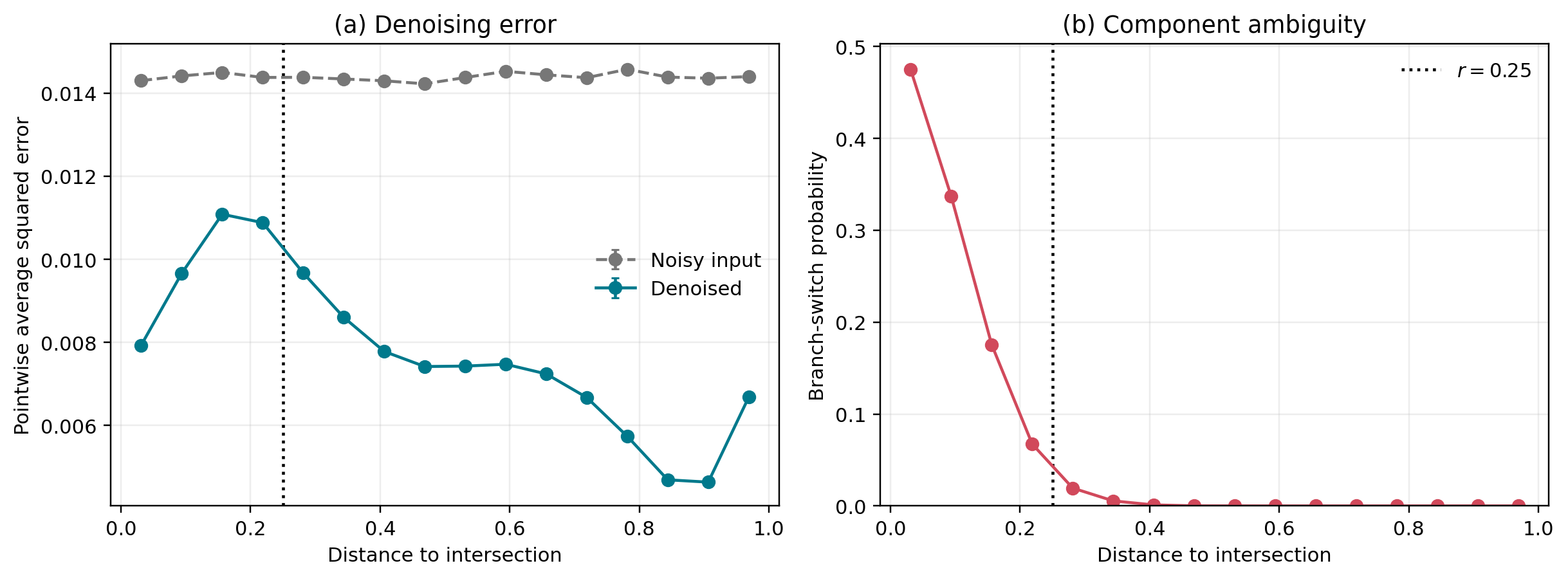}
    \caption{Denoising behavior as a function of distance from the intersection for the X-shaped union. Left: squared errors of the noisy inputs and denoised outputs. Right: component-switch probability. Each quantity is averaged over clean test samples and $64$ independent noise realizations. The dotted line marks the intersection-neighborhood radius $r=0.25$.}
    \label{fig:xshape_distance}
\end{figure}

Figure~\ref{fig:xshape_distance} examines how the denoising behavior varies with the distance from the intersection for the X-shaped union. The noisy-input error is approximately constant across distances, as expected from the fixed Gaussian noise level. 
In contrast, the denoised error depends strongly on the distance to the intersection: it increases near the crossing, reaches its largest values shortly outside the intersection point, and then decreases as the two branches become well separated. The component-switch probability is largest near the crossing and rapidly decays to nearly zero in the non-intersection region. This behavior is consistent with the ambiguity interpretation: at the exact crossing the two components share the same point, while slightly away from the crossing the two branches are distinct but still difficult to identify from noisy observations.

\begin{figure}[!htbp]
    \centering
    \includegraphics[width=0.9\linewidth]{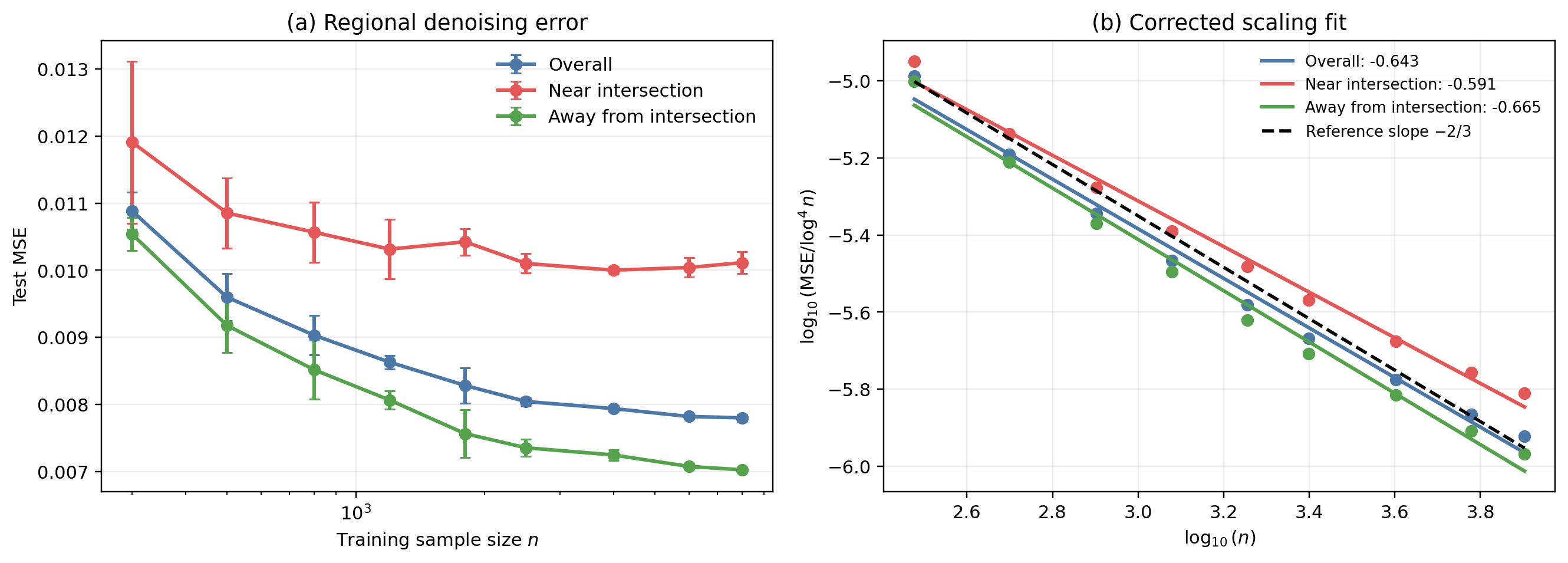}
    \caption{Sample-size scaling for the X-shaped union in the intersection neighborhood and the non-intersection region. Left: test MSE versus training sample size. Right: corrected log--log fits of $\mathrm{MSE}/\log^4 n$. The fitted slopes are $-0.643$ overall, $-0.591$ in the intersection neighborhood, and $-0.665$ in the non-intersection region; the dashed reference line corresponds to the theoretical slope $\alpha_{\mathrm{theory}} = -2/(d+2)=-2/3$.}
    \label{fig:xshape_near_away_scaling}
\end{figure}

We next compare the sample-size behavior in the intersection neighborhood and the non-intersection region. As shown in Figure~\ref{fig:xshape_near_away_scaling}, the intersection-neighborhood error remains larger and approaches a higher empirical floor. After correcting for the $\log^4 n$ factor, the fitted slopes are $-0.591$ in the intersection neighborhood and $-0.665$ in the non-intersection region. The latter is close to the single-manifold prediction
$\alpha_{\mathrm{theory}}=-2/(d+2)=-2/3$, while the less negative intersection-neighborhood slope reflects slower
error decay near the intersection. The overall empirical slope is $-0.643$.

\begin{table}[H]
\centering
\small
\renewcommand{\arraystretch}{1.15}
\begin{tabular}{lccccccc}
\toprule
Geometry
& $(D,d)$
& $\dim(\cI)$
& $\sigma$
& $\alpha_{\mathrm{theory}}$
& $\hat{\alpha}_{\mathrm{overall}}$
& $\hat{\alpha}_{\mathrm{int}}$
& $\hat{\alpha}_{\mathrm{non}}$ \\
\midrule
X-shaped lines
& $(2,1)$ & 0 & 0.12 & -0.667 & -0.643 & -0.591 & -0.665 \\
Two intersecting circles
& $(2,1)$ & 0 & 0.12 & -0.667 & -0.679 & -0.607 & -0.694 \\
Two intersecting $2$-spheres
& $(3,2)$ & 1 & 0.12 & -0.500 & -0.586 & -0.552 & -0.595 \\
\bottomrule
\end{tabular}
\caption{Empirical log--log slopes for the intersecting synthetic examples. Here $D$ and $d$ denote the ambient and component dimensions, respectively, and $\dim(\cI)$ denotes the dimension of the intersection set. The slopes are estimated from corrected log--log fits of $\mathrm{MSE}/\log^4 n$, reported overall, in the intersection neighborhood, and in the non-intersection region.}
\label{tab:intersecting_scaling}
\end{table}

Table~\ref{tab:intersecting_scaling} summarizes the empirical log--log slopes across the three intersecting synthetic geometries. In each case, the slope in the intersection neighborhood is less negative than that in the non-intersection region, indicating slower error decay near the intersections. The non-intersection slopes remain close to the single-manifold prediction, consistent with Theorem~\ref{thm:intersecting_union}, which predicts an additional ambiguity contribution from the intersection neighborhood.

\subsection{Image datasets}
\label{sec:image_datasets}

We next consider three real image datasets: MNIST, Extended Yale B, and OCTMNIST. MNIST consists of grayscale handwritten-digit images, Extended Yale B contains face images under varying illumination conditions, and OCTMNIST contains retinal optical coherence tomography images from four diagnostic categories: choroidal neovascularization (CNV), diabetic macular edema (DME), drusen, and normal retina. Since representative MNIST denoising examples are already shown in Figure~\ref{fig:mnist_dae_demo}, we focus here on Extended Yale B and OCTMNIST.

\begin{figure}[!htbp]
\centering 
\includegraphics[width=0.9\linewidth]{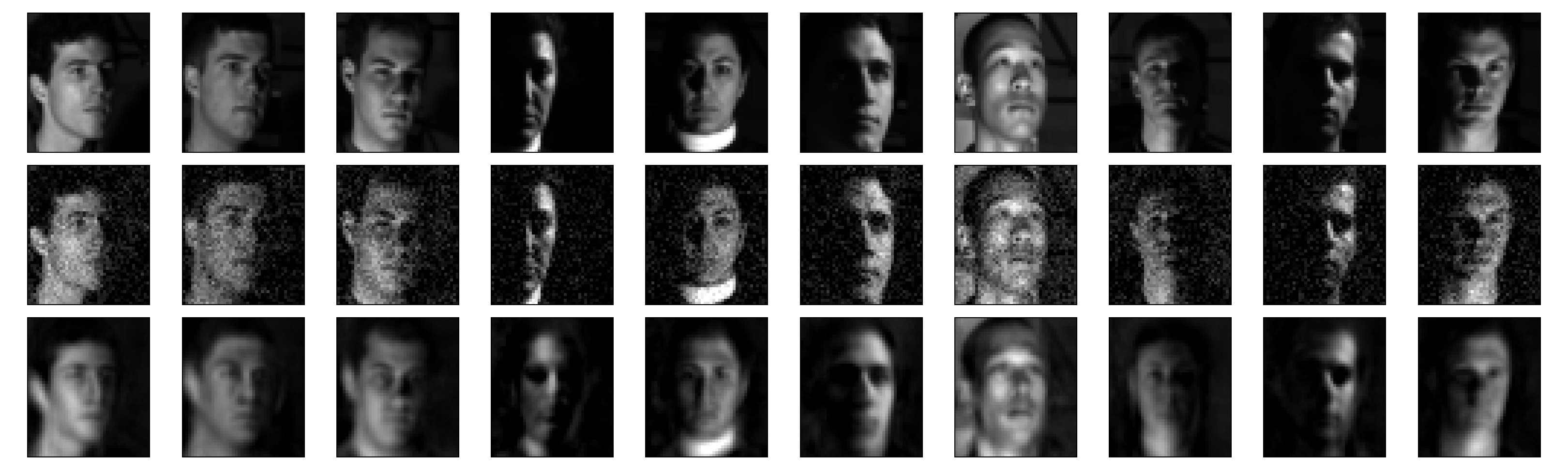} \caption{Representative denoising examples on Extended Yale B. Columns show different test images, while the three rows correspond to clean images, noisy inputs, and denoised reconstructions.} \label{fig:yaleb_demo} 
\end{figure}

\begin{figure}[!htbp]
\centering 
\includegraphics[width=0.9\linewidth]{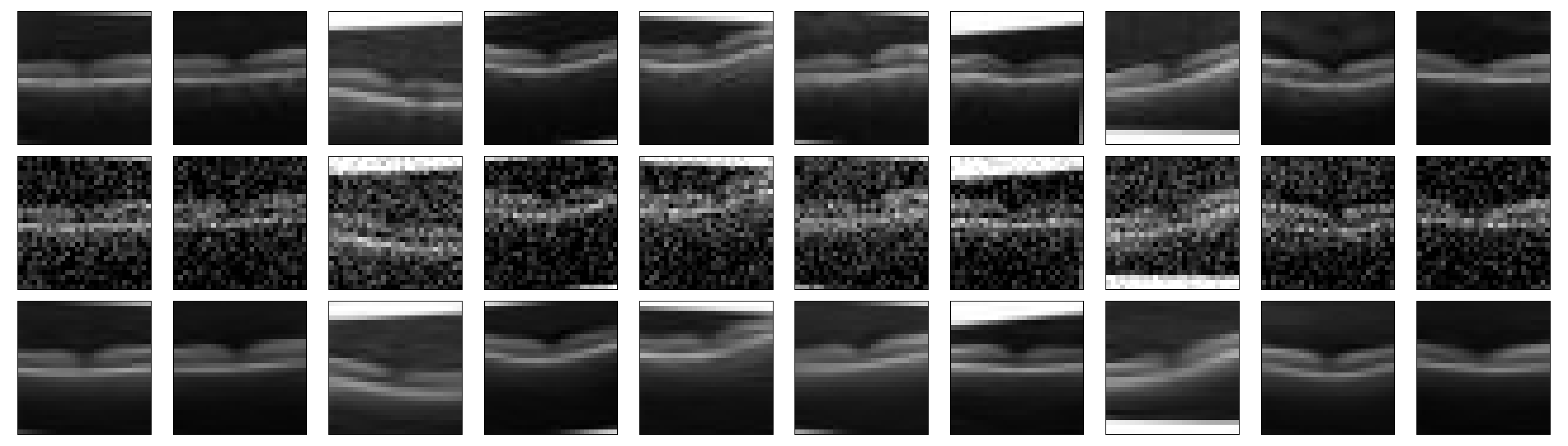} \caption{Representative denoising examples on OCTMNIST. Columns show test images from different retinal OCT classes, while the rows correspond to clean images, noisy inputs, and denoised reconstructions.} 
\label{fig:octmnist_demo} 
\end{figure}

Figures~\ref{fig:yaleb_demo} and~\ref{fig:octmnist_demo} show representative denoising results on Extended Yale B and OCTMNIST, respectively. In both cases, the denoised reconstructions remove substantial pixel-level noise while preserving the main image structures.

\begin{figure}[!htbp]
    \centering
    \includegraphics[width=0.98\linewidth]
        {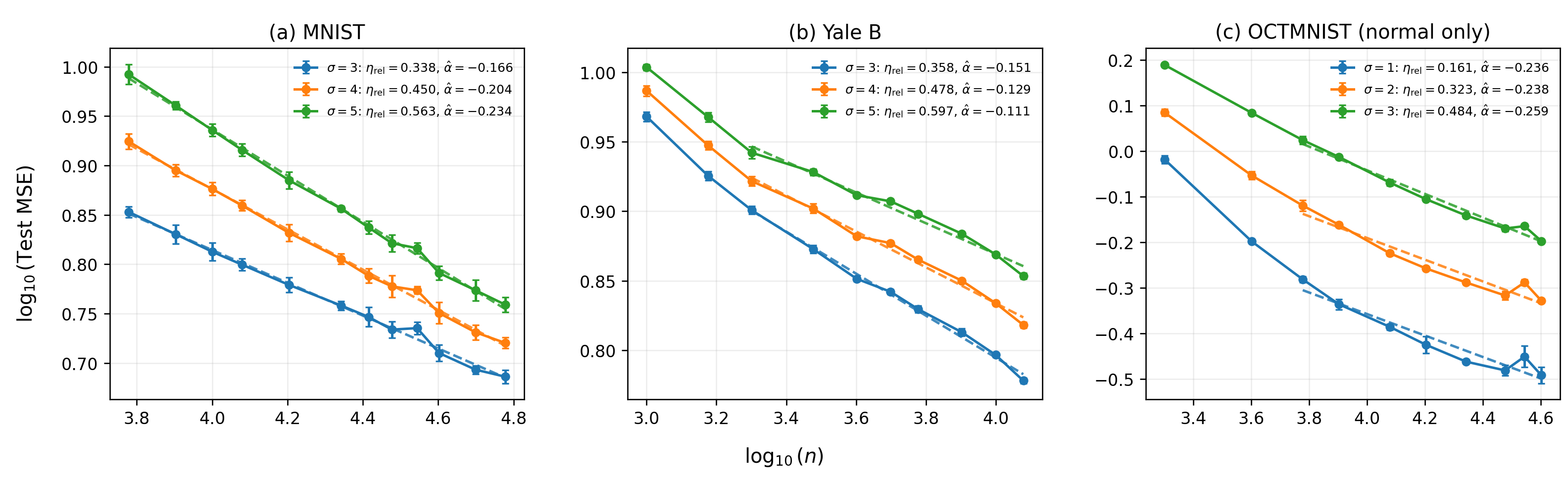}
    \caption{Sample-size scaling of the test denoising error on real
    image datasets. Each curve shows the test MSE as a function of the
    training sample size $n$ on a log--log scale, and the dashed lines
    show the fitted trends used to estimate the empirical log--log slopes.
    The MNIST fits use all sample sizes ($n=6000$--$60000$); the Extended
    Yale B and normal-only OCTMNIST fits use $n\ge 2000$ and $n\ge 6000$,
    respectively. The legend slopes and Table~\ref{tab:image_scaling}
    are calculated from the same fits.}
    \label{fig:image_scaling}
\end{figure}

Figure~\ref{fig:image_scaling} reports the sample-size scaling of the test denoising error on MNIST, Extended Yale B, and OCTMNIST normal images. Since the underlying geometry of real image datasets is unknown, the noise level cannot be calibrated directly by the reach or tubular-neighborhood radius used in the theory. Therefore, we report the relative noise level $\eta_{\mathrm{rel}} = \frac{\|x_{\mathrm{noisy}}-x_{\mathrm{clean}}\|_2}{\|x_{\mathrm{clean}}\|_2}$ averaged over the dataset. This quantity measures the perturbation magnitude relative to the signal energy of the clean image. 

For each dataset and noise level, we estimate the empirical log--log slope $\hat{\alpha}_{\mathrm{empirical}}$ from the decay of the test MSE with the training sample size. The theory predicts $\alpha_{\mathrm{theory}} =
-\frac{2}{d+2}.$ If the empirical test error behaves approximately as $n^{\hat{\alpha}_{\mathrm{empirical}}}$, we define the corresponding heuristic effective dimension by
$d_{\mathrm{eff}} = -\frac{2}{\hat{\alpha}_{\mathrm{empirical}}}-2$.

Table~\ref{tab:image_scaling} summarizes the empirical sample-size rates and the corresponding heuristic effective dimensions. Across all three datasets, the inferred values of $d_{\mathrm{eff}}$ are substantially smaller than the ambient dimension $D$, consistent with the presence of low-dimensional structure in the image data. For OCTMNIST, the scaling experiment is restricted to the normal class in both training and testing. As an additional comparison, a local maximum-likelihood intrinsic-dimension estimator applied to the clean normal OCTMNIST images gives estimates of approximately $10$--$11.5$ across several choices of the neighborhood size $k$, where $k$ denotes the number of nearest neighbors used by the estimator. These estimates provide additional evidence of low-dimensional structure, although they are not expected to coincide with the scaling-based quantity $d_{\mathrm{eff}}$.

\begin{table}[H]
\centering
\small
\renewcommand{\arraystretch}{1.15}
\begin{tabular}{lccccc}
\toprule
Dataset & $\sigma$ & $D$ & Mean relative noise level & $\hat{\alpha}_{\mathrm{empirical}}$ & $d_{\mathrm{eff}}$ \\
\midrule
MNIST & 3 & 784 & 0.338 & -0.166 & 10.0 \\
MNIST & 4 & 784 & 0.450 & -0.204 & 7.8 \\
MNIST & 5 & 784 & 0.563 & -0.234 & 6.6 \\
Yale B & 3 & 2016 & 0.358 & -0.151 & 11.2 \\
Yale B & 4 & 2016 & 0.478 & -0.129 & 13.5 \\
Yale B & 5 & 2016 & 0.597 & -0.111 & 16.1 \\
OCTMNIST (normal only) & 1 & 784 & 0.161 & -0.236 & 6.5 \\
OCTMNIST (normal only) & 2 & 784 & 0.323 & -0.238 & 6.4 \\
OCTMNIST (normal only) & 3 & 784 & 0.484 & -0.259 & 5.7 \\
\bottomrule
\end{tabular}
\caption{Empirical log--log slopes and corresponding heuristic effective dimensions for the real image datasets. The mean relative noise level is the dataset average of $\|x_{\mathrm{noisy}}-x_{\mathrm{clean}}\|_2/\|x_{\mathrm{clean}}\|_2$.}
\label{tab:image_scaling}
\end{table}

\begin{figure}[t]
    \centering
    \includegraphics[width=0.90\linewidth]
        {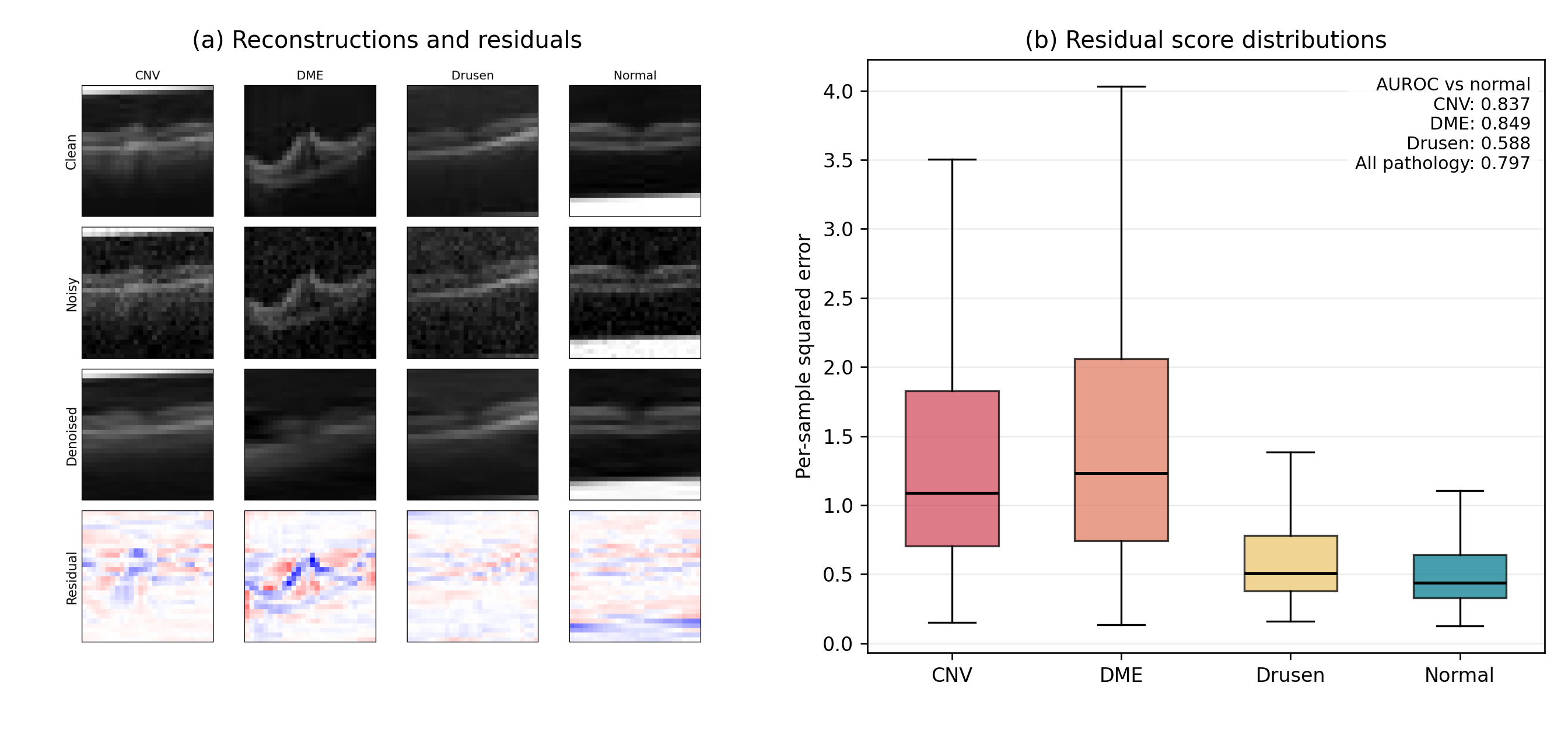}
    \caption{Residual-based anomaly analysis on OCTMNIST using a
    denoising autoencoder trained on $4000$ normal OCT images.
    Left: representative clean images, noisy observations, denoised
    reconstructions, and residuals $\widehat{\xb}-\xb$ for CNV, DME,
    drusen, and normal retina.
    Right: class-wise distributions of the squared reconstruction error;
    the annotation reports AUROC values for distinguishing each
    pathological class, and all pathological classes jointly, from the
    normal class.}
    \label{fig:octmnist_anomaly}
\end{figure}

Finally, we provide an additional anomaly-detection interpretation using OCTMNIST. The denoising autoencoder is trained only on normal OCT images and then evaluated on normal images and the three pathological classes: choroidal neovascularization (CNV), diabetic macular edema (DME), and drusen. As shown in Figure~\ref{fig:octmnist_anomaly}, normal images, which match the training distribution, are reconstructed more accurately, whereas pathological images tend to produce larger reconstruction residuals. We quantify this separation using the area under the receiver operating characteristic curve (AUROC), with the squared reconstruction error used as the anomaly score. Let
$s(\xb) = \|\widehat{\cG}(\yb)-\xb\|_2^2$ denote the anomaly score. Then the AUROC can be written as
$\mathrm{AUROC} = \PP\!\left(s(\xb_{\mathrm{path}}) > s(\xb_{\mathrm{normal}}) \right)$, where $\xb_{\mathrm{path}}$ and $\xb_{\mathrm{normal}}$ are independently sampled pathological and normal images, respectively. Thus, $\mathrm{AUROC}=0.5$ corresponds to random discrimination, while larger values indicate better separation.

CNV and DME exhibit substantially larger reconstruction errors than normal images, with AUROC values of $0.837$ and $0.849$, respectively, while drusen overlaps more strongly with the normal distribution, yielding an AUROC of $0.588$. Combining all three pathological classes gives an AUROC of $0.797$.

\section{Proofs of the main results}
\label{sec:proofs}

In this section, we prove the main results for the three geometric scenarios considered in Section~\ref{sec:main_results}. We first establish the single-manifold result, then extend the argument to a finite union of pairwise separated manifolds, and finally treat the intersecting-manifold case.

For notational convenience, given encoder and decoder classes $\cF_{\mathrm{NN}}^{\cE}$ and $\cF_{\mathrm{NN}}^{\cD}$, define $ \cF_{\mathrm{NN}}^{\cG} = \{\cD\circ\cE:\cE\in\cF_{\mathrm{NN}}^{\cE},
\cD\in\cF_{\mathrm{NN}}^{\cD}\}$, and write $\widehat{\cG} = \widehat{\cD}\circ\widehat{\cE}$, where $(\widehat{\cE},\widehat{\cD})$ is a global minimizer of the empirical reconstruction loss in \eqref{eq:dae_training}.

\subsection{Proof of Theorem~\ref{thm:single_manifold}}
\label{proof:single_manifold}
\begin{proof}[Proof of Theorem~\ref{thm:single_manifold}]
The proof follows the five-step strategy outlined in Section~\ref{sec:single_manifold}. We first control a high-probability projection event, then decompose the expected reconstruction error into a bias term and a variance term. The bias term is further split into an oracle approximation term and a geometric projection term. The oracle approximation is controlled by Lemma~\ref{lem:multichart_oracle_approx}, and the variance term is controlled by a covering-number argument. Finally, we combine these bounds and balance the approximation accuracy with the network size.

\noindent
{\bf Step 1: Controlling a high-probability projection event.}
Recall the projection-valid event $E_q$ defined in \eqref{eq:projection_valid_event}. We decompose the ambient Gaussian noise into its tangential and normal components,

$\bepsilon=\bepsilon_T+\bepsilon_N$, where $\bepsilon_T=\operatorname{Proj}_{T_{\xb}\cM}(\bepsilon)$ and $\bepsilon_N=\operatorname{Proj}_{T_{\xb}^{\perp}\cM}(\bepsilon)$.

\begin{lemma}[High-probability tubular event]
\label{lem:tubular_event}
Under Setting~\ref{setting:data} and Scenario~\ref{scenario:single},
if $q\tau_{\cM}>\sigma$, then
$
\PP(E_q^c)
\le
\exp\!\left[
-\frac D4\left(
\sqrt{\frac{2q^2\tau_{\cM}^2}{\sigma^2}-1}-1
\right)^2
\right].
$
Moreover, on the event $E_q$, the orthogonal projection
$\pi(\yb)$ onto $\cM$ is uniquely defined. In addition,
$
\EE\!\left[
\|\bepsilon_T\|_2^2
\mid
\xb
\right]
=
\frac{d\sigma^2}{D}.
$
\end{lemma}
The proof of Lemma~\ref{lem:tubular_event} is given in
Appendix~\ref{app:tubular_event}. Lemma~\ref{lem:tubular_event}
shows that the global event $E_q$ occurs with high probability when the noise level $\sigma$ is small relative to the geometric scale $q\tau_{\cM}$.

\noindent
\textbf{Step 2: Error decomposition.}
Define the population and empirical reconstruction risks by $L(\cG) := \EE_{(\yb,\xb)\sim\gamma}
\|\cG(\yb)-\xb\|_2^2$ and $L_{\cS}(\cG) := \frac{1}{n}\sum_{i=1}^n \|\cG(\yb_i)-\xb_i\|_2^2$. Adding and subtracting $2L_{\cS}(\widehat{\cG})$, we decompose
\begin{equation}
\label{eq:risk_decomposition}
\EE_{\cS}L(\widehat{\cG})
=
T_1+T_2,
\end{equation}
where $T_1 = 2\,\EE_{\cS}L_{\cS}(\widehat{\cG})$ and $T_2 = \EE_{\cS} \left[L(\widehat{\cG}) - 2L_{\cS}(\widehat{\cG}) \right]$. Here, $T_1$ is controlled through neural-network approximation of the projection-based oracle map, while $T_2$ is the stochastic term controlled by the complexity of the network class.

\noindent
{\bf Step 3: Bounding the bias term.}
Since $\widehat{\cG}$ minimizes $L_{\cS}$ over $\cF_{\mathrm{NN}}^{\cG}$, we have
$$
T_1
=
2\,\EE_{\cS}\!\left[
\inf_{\cG\in\cF_{\mathrm{NN}}^{\cG}}L_{\cS}(\cG)
\right]
\le
2\inf_{\cG\in\cF_{\mathrm{NN}}^{\cG}}
\EE_{\cS}L_{\cS}(\cG)
=
2\inf_{\cG\in\cF_{\mathrm{NN}}^{\cG}}L(\cG),
$$
where the last identity uses the i.i.d.\ assumption.

For any event $A$, let $\mathbf{1}_A$ denote its indicator function. For any $\cG\in\cF_{\mathrm{NN}}^{\cG}$, taking expectation with respect to $(\yb,\xb)\sim\gamma$, we decompose
$$
L(\cG)
=
\EE\!\left[
\norm{\cG(\yb)-\xb}_2^2\mathbf{1}_{E_q}
\right]
+
\EE\!\left[
\norm{\cG(\yb)-\xb}_2^2\mathbf{1}_{E_q^c}
\right].
$$
On the event $E_q$, the projection $\pi(\yb)$ is well defined. Hence
$$
\norm{\cG(\yb)-\xb}_2^2
\le
2\norm{\cG(\yb)-\pi(\yb)}_2^2
+
2\norm{\pi(\yb)-\xb}_2^2.
$$
Therefore,
$$
L(\cG)
\le
2\EE\!\left[\norm{\cG(\yb)-\pi(\yb)}_2^2\,\mathbf 1_{E_q}\right]
+
2\EE\!\left[\norm{\pi(\yb)-\xb}_2^2\,\mathbf 1_{E_q}\right]
+
\EE\!\left[\norm{\cG(\yb)-\xb}_2^2\,\mathbf 1_{E_q^c}\right].
$$
Since $\xb\in\cM\subset[-B,B]^D$ and the network outputs are bounded in $[-B,B]^D$, we have $\norm{\cG(\yb)-\xb}_2^2\le 4DB^2$. Hence
$$
L(\cG)
\le
2\EE\!\left[\norm{\cG(\yb)-\pi(\yb)}_2^2\,\mathbf 1_{E_q}\right]
+
2\EE\!\left[\norm{\pi(\yb)-\xb}_2^2\,\mathbf 1_{E_q}\right]
+
4DB^2\,\PP(E_q^c).
$$
Using Lemma~\ref{lem:tubular_event} and
$T_1\le 2\inf_{\cG\in\cF_{\mathrm{NN}}^{\cG}}L(\cG)$,
we obtain
\begin{align*}
T_1
&\le
4\inf_{\cG\in\cF_{\mathrm{NN}}^{\cG}}
\EE\!\left[
\norm{\cG(\yb)-\pi(\yb)}_2^2\mathbf 1_{E_q}
\right]
+
4\EE\!\left[
\norm{\pi(\yb)-\xb}_2^2\mathbf 1_{E_q}
\right]\\
&\quad+
8DB^2
\exp\!\left[
-\frac D4\left(
\sqrt{\frac{2q^2\tau_{\cM}^2}{\sigma^2}-1}-1
\right)^2
\right].
\end{align*}
The first term is controlled through a multi-chart oracle encoder--decoder construction based on the manifold projection, following the framework in \cite{liu2024chart}.

\begin{lemma}[Multi-chart oracle construction, adapted from
{\cite[Appendix~A.4]{liu2024chart}}]
\label{lem:multichart_oracle}
Under Setting~\ref{setting:data} and Scenario~\ref{scenario:single},
fix $q\in(0,1)$ and consider the finite atlas
$\{(U_j,\phi_j)\}_{j=1}^{C_{\cM}}$
from Lemma~\ref{lem:multichart_construction}.
There exist a partition of unity on $\cM(q)$, an encoder $\cE:\cM(q)\to\RR^{C_{\cM}(d+1)}$, and a decoder $\cD:\RR^{C_{\cM}(d+1)}\to\RR^D$ such that for any $\yb\in\cM(q)$, $\cD\circ\cE(\yb)=\pi(\yb)$.
\end{lemma}
The proof of Lemma \ref{lem:multichart_oracle} is given in Appendix~\ref{app:multichart_oracle}. Lemma \ref{lem:multichart_oracle} constructs a pair of oracle encoder and decoder. We next construct neural networks to approximate this oracle encoder and decoder.

\begin{lemma}[Neural approximation of the multi-chart oracle, adapted from {\cite[Lemma~11]{liu2024chart}}]
\label{lem:multichart_oracle_approx}
Under Setting~\ref{setting:data} and Scenario~\ref{scenario:single},
fix $q\in(0,1)$ and consider the oracle encoder--decoder pair in
Lemma~\ref{lem:multichart_oracle}. For any $0<\varepsilon<1$, there exist
networks $\widetilde{\cE}\in\cF_{\mathrm{NN}}^{\cE}$ and
$\widetilde{\cD}\in\cF_{\mathrm{NN}}^{\cD}$, where
$\cF_{\mathrm{NN}}^{\cE}=\cF(D,C_{\cM}(d+1);L_{\cE},p_{\cE},K_{\cE},\kappa_{\cE},R_{\cE})$,
with $L_{\cE}=O(\log^2\varepsilon^{-1}+\log D)$,
$p_{\cE}=O(D\varepsilon^{-d})$,
$K_{\cE}=O((D\log D)\varepsilon^{-d}\log^2\varepsilon^{-1})$,
$\kappa_{\cE}=O(\varepsilon^{-2})$, and
$R_{\cE}=\max\{\tau_{\cM}/4,1\}$, and
$\cF_{\mathrm{NN}}^{\cD}=\cF(C_{\cM}(d+1),D;L_{\cD},p_{\cD},K_{\cD},\kappa_{\cD},R_{\cD})$,
with $L_{\cD}=O(\log^2\varepsilon^{-1}+\log D)$,
$p_{\cD}=O(D\varepsilon^{-d})$,
$K_{\cD}=O(D\varepsilon^{-d}\log^2\varepsilon^{-1}+D\log D)$,
$\kappa_{\cD}=O(\varepsilon^{-1})$, and $R_{\cD}=B$. Then
$$
\textstyle\sup_{\yb\in\cM(q)}
\left\|
\widetilde{\cD}\circ\widetilde{\cE}(\yb)-\pi(\yb)
\right\|_\infty
\le
\varepsilon.
$$
The constants hidden in the $O(\cdot)$ notation depend on
$d,\tau_{\cM},q,B,C_{\cM}$, and the volume of $\cM$.
\end{lemma}

The proof of Lemma~\ref{lem:multichart_oracle_approx} is given in
Appendix~\ref{app:multichart_oracle_approx}. The main idea is to approximate the local chart maps and partition-of-unity functions by ReLU networks and then combine these local approximations according to the oracle construction in Lemma~\ref{lem:multichart_oracle}. 

Using Lemma~\ref{lem:multichart_oracle_approx}, we continue bounding the first term in $T_1$. Let
$\widetilde{\cG} := \widetilde{\cD}\circ\widetilde{\cE}$, where $\widetilde{\cE}$ and $\widetilde{\cD}$ are the networks constructed in Lemma~\ref{lem:multichart_oracle_approx} with approximation accuracy $\varepsilon$. On $E_q$, $\operatorname{dist}(\yb,\cM)
\le \|\bepsilon\|_2 < q\tau_{\cM}$. Hence $\pi(\yb)$ is uniquely defined, with $\yb-\pi(\yb)\in T_{\pi(\yb)}^\perp\cM$, and
$\|\yb-\pi(\yb)\|_2 < q\tau_{\cM} \le q\tau_{\cM}(\pi(\yb))$. Therefore $\yb\in\cM(q)$. We then have
$$
\inf_{\cG\in\cF_{\mathrm{NN}}^{\cG}}
\EE\!\left[
\norm{\cG(\yb)-\pi(\yb)}_2^2\mathbf{1}_{E_q}
\right]
\le
\EE\!\left[
\norm{\widetilde{\cG}(\yb)-\pi(\yb)}_2^2\mathbf{1}_{E_q}
\right]
\le
D\varepsilon^2.
$$ 
We next bound the geometric projection term. On the event $E_q$, $
\norm{\bepsilon_N}_2 \le \norm{\bepsilon}_2 < q\tau_{\cM} \le q\tau_{\cM}(\xb)$. Hence $\xb+\bepsilon_N$ lies in the projection-valid tubular neighborhood of $\cM$ and satisfies $\pi(\xb+\bepsilon_N)=\xb$. Since $\yb = \xb+\bepsilon_N+\bepsilon_T$, we have $\pi(\yb)-\xb = \pi(\xb+\bepsilon_N+\bepsilon_T) - \pi(\xb+\bepsilon_N)$. Using the Lipschitz continuity of the nearest-point projection on $\cM(q)$ from Lemma~2.1 of \cite{cloninger2021deep}, we obtain $\norm{\pi(\yb)-\xb}_2 \le (1-q)^{-1}\norm{\bepsilon_T}_2$. Therefore, by Lemma~\ref{lem:tubular_event},
$$
\EE\!\left[
\norm{\pi(\yb)-\xb}_2^2\mathbf{1}_{E_q}
\right]
\le
(1-q)^{-2}
\EE\!\left[
\norm{\bepsilon_T}_2^2\mathbf{1}_{E_q}
\right]
\le
(1-q)^{-2}
\frac{d\sigma^2}{D}.
$$
Substituting these estimates into the bound for $T_1$, we obtain
\begin{equation}
\label{eq:T1_bound}
\begin{aligned}
T_1
\le\;
4D\varepsilon^2
+
4(1-q)^{-2}\frac{d\sigma^2}{D}
+
8DB^2
\exp\!\left[
-\frac D4\left(
\sqrt{\frac{2q^2\tau_{\cM}^2}{\sigma^2}-1}-1
\right)^2
\right].
\end{aligned}
\end{equation}

\noindent
{\bf Step 4: Bounding the variance term.}

Since the Gaussian noisy observation $\yb$ is not globally bounded,
we first separate the contribution from the good event $E_q$ and its
complement. For any $\cG\in\cF_{\mathrm{NN}}^{\cG}$, write
$$
\|\cG(\yb)-\xb\|_2^2
=
\|\cG(\yb)-\xb\|_2^2\mathbf 1_{E_q}
+
\|\cG(\yb)-\xb\|_2^2\mathbf 1_{E_q^c}.
$$
Using this decomposition in the definition of $T_2$, we obtain
\begin{align*}
T_2
={}&
\EE_{\cS}\Bigg[
\EE\!\left[
\|\widehat{\cG}(\yb)-\xb\|_2^2\mathbf 1_{E_q}
\right]
-
\frac{2}{n}\sum_{i=1}^n
\|\widehat{\cG}(\yb_i)-\xb_i\|_2^2
\mathbf 1_{E_{q,i}}
\Bigg]
\\
&+
\EE_{\cS}\Bigg[
\EE\!\left[
\|\widehat{\cG}(\yb)-\xb\|_2^2\mathbf 1_{E_q^c}
\right]
-
\frac{2}{n}\sum_{i=1}^n
\|\widehat{\cG}(\yb_i)-\xb_i\|_2^2
\mathbf 1_{E_{q,i}^c}
\Bigg],
\end{align*}
where $E_{q,i}:=\{\|\bepsilon_i\|_2<q\tau_{\cM}\}$. Since the empirical term in the second bracket is nonpositive and $\|\cG(\yb)-\xb\|_2^2\le 4DB^2$, we have
\begin{align*}
T_2
\le{}&
\EE_{\cS}\Bigg[
\EE\!\left[
\|\widehat{\cG}(\yb)-\xb\|_2^2\mathbf 1_{E_q}
\right]
-
\frac{2}{n}\sum_{i=1}^n
\|\widehat{\cG}(\yb_i)-\xb_i\|_2^2
\mathbf 1_{E_{q,i}}
\Bigg]
+
4DB^2\,\PP(E_q^c).
\end{align*}
It remains to control the first term, which only involves samples on
the good event $E_q$. On $E_q$, $\|\yb\|_\infty \le \|\xb\|_\infty+\|\bepsilon\|_2 < B+q\tau_{\cM}$. Define $B_q:=B+q\tau_{\cM}$. Thus the noisy input is restricted to the bounded domain $[-B_q,B_q]^D$ on $E_q$.

We now define the good-event loss class $ \cH_q := \left\{(\yb,\xb)\mapsto \|\cG(\yb)-\xb\|_2^2\mathbf 1_{E_q} : \cG\in\cF_{\mathrm{NN}}^{\cG} \right\}$. The same symmetrization and metric-entropy argument as in
Lemma~2 of \cite{liu2024chart} can therefore be applied to
$\cH_q$ on the bounded input domain $[-B_q,B_q]^D$.
\begin{lemma}[Variance bound on a good event]
\label{lem:variance_bound}
Let $E$ be an event such that $\|\yb\|_\infty\le B_E$ on $E$, for some $B_E>0$. Then, for any $0<\eta<1$,
\begin{align*}
T_2
\le\;&
\frac{35DB^2}{n}
\log N\!\left(
\frac{\eta}{4DB},
\cF_{\mathrm{NN}}^{\cG},
\|\cdot\|_{L^{\infty,\infty}([-B_E,B_E]^D)}
\right)
+
6\eta
+
4DB^2\,\PP(E^c).
\end{align*}
\end{lemma}

By the network-composition and covering-number argument in
Lemma~3 of \cite{liu2024chart},
\begin{align*}
&N\!\left(
\delta,
\cF_{\mathrm{NN}}^{\cG},
\|\cdot\|_{L^{\infty,\infty}([-B_q,B_q]^D)}
\right) \le
\\
&\left(
\frac{
2(L_{\cE}+L_{\cD})^2
\bigl(\max\{p_{\cE},p_{\cD}\}B_q+2\bigr)
\bigl(\max\{\kappa_{\cE},\kappa_{\cD}\}\bigr)^{L_{\cE}+L_{\cD}}
\bigl(\max\{p_{\cE},p_{\cD}\}\bigr)^{L_{\cE}+L_{\cD}+1}
}{
\delta
}
\right)^{2(K_{\cE}+K_{\cD})}
\end{align*}
Since $B_q=B+q\tau_{\cM}$ is fixed with respect to
$n$ and $\varepsilon$, substituting the network-size bounds from Lemma~\ref{lem:multichart_oracle_approx} yields
\begin{align*}
&\log N\!\left(
\frac{\eta}{4DB},
\cF_{\mathrm{NN}}^{\cG},
\|\cdot\|_{L^{\infty,\infty}([-B_q,B_q]^D)}
\right)
\\
& \le O\!\left(
(D\log^3 D)\varepsilon^{-d}\log^4\varepsilon^{-1}
+
(D\log^3 D)\varepsilon^{-d}
\log^3\varepsilon^{-1}\log\eta^{-1}
\right).
\end{align*}
Therefore, by Lemma~\ref{lem:tubular_event},
\begin{equation}
\label{eq:T2_bound}
\begin{aligned}
T_2
\le\;&
C(D^2\log^3 D)\,
n^{-1}\varepsilon^{-d}\log^4\varepsilon^{-1}
+
C(D^2\log^3 D)\,
n^{-1}\varepsilon^{-d}
\log^3\varepsilon^{-1}\log\eta^{-1} \\
&+
6\eta
+
4DB^2
\exp\!\left[
-\frac D4\left(
\sqrt{\frac{2q^2\tau_{\cM}^2}{\sigma^2}-1}-1
\right)^2
\right].
\end{aligned}
\end{equation}

\noindent
\textbf{Step 5: Combining the bounds.}
Combining \eqref{eq:T1_bound} and \eqref{eq:T2_bound} and choosing $\varepsilon=n^{-1/(d+2)}$ and $\eta=n^{-1}$, we obtain \eqref{eq:errorsingle}, completing the proof.

\end{proof}

\subsection{Proof of Theorem~\ref{thm:union_manifold}}
\label{proof:union_manifold}

\begin{proof}[Proof of Theorem~\ref{thm:union_manifold}]
The proof follows the same five-step strategy as the proof of
Theorem~\ref{thm:single_manifold}. The main additional ingredient is
stable component identification, which allows the single-manifold
projection and approximation arguments to be applied componentwise.

\noindent
\textbf{Step 1: Controlling the projection-valid and component-identification event.}
Define $E_{q,*} := \left\{\|\bepsilon\|_2<q\tau_*\right\}$ and $\tau_*:=\min_{1\le \ell\le L}\tau_{\cM_\ell}$. Since $q<1$ and $\Delta>\tau_*$, we have $q\tau_*<\Delta$. Thus, on $E_{q,*}$, a noisy observation remains within the projection-valid neighborhood of its source component and cannot become closer to a different component.

\begin{lemma}[High-probability tubular event on a union of manifolds]
\label{lem:union_tubular_event}

Under Setting~\ref{setting:data} and Scenario~\ref{scenario:separated},
if $q\tau_*>\sigma$,
$
\PP(E_{q,*}^c)
\le
\exp\!\left[
-\frac D4\left(
\sqrt{\frac{2q^2\tau_*^2}{\sigma^2}-1}-1
\right)^2
\right].
$
Moreover, on $E_{q,*}$, the nearest-point projection of
$\yb=\xb+\bepsilon$ onto $\cM=\bigcup_{\ell=1}^L\cM_\ell$ is uniquely defined and belongs to the same component manifold as
$\xb$.
\end{lemma}

The proof is given in Appendix~\ref{app:union_projection}.

\noindent
\textbf{Step 2: Error decomposition.}

We use the same population risk $L$, empirical risk $L_{\cS}$, and
decomposition $\EE_{\cS}L(\widehat{\cG}) = T_1+T_2$ as in the proof of Theorem~\ref{thm:single_manifold}.

\noindent
\textbf{Step 3: Bounding the bias term.}
As in the proof of Theorem~\ref{thm:single_manifold},
the global ERM property gives $T_1\le 2\inf_{\cG\in\cF_{\mathrm{NN}}^{\cG}}L(\cG)$.
The key difference is that the separated union $\cM=\bigcup_{\ell=1}^L\cM_\ell$ is a compact, possibly disconnected, smooth $d$-dimensional manifold with $\tau_{\cM}\ge\min\{\tau_*,\Delta\}=\tau_*$. Thus, the multi-chart oracle construction and
neural approximation in Lemmas~\ref{lem:multichart_oracle} and~\ref{lem:multichart_oracle_approx} apply componentwise using the combined atlas with $C_*:=\sum_{\ell=1}^L C_{\cM_\ell}$. Consequently, there exists $\widetilde{\cG}\in\cF_{\mathrm{NN}}^{\cG}$ with the network-size scaling in Theorem~\ref{thm:union_manifold} such that $\|\widetilde{\cG}(\yb)-\pi(\yb)\|_\infty \le\varepsilon$ on $E_{q,*}$.

For $\xb\in\cM_\ell$, decompose $\bepsilon=\bepsilon_T+\bepsilon_N$ relative to $T_{\xb}\cM_\ell$. By Lemma~\ref{lem:union_tubular_event}, the projection remains on the correct component on $E_{q,*}$, so $\pi(\yb)=\pi_\ell(\yb)$ and $\pi_\ell(\xb+\bepsilon_N)=\xb$. The projection Lipschitz bound therefore gives $\|\pi(\yb)-\xb\|_2
\le(1-q)^{-1}\|\bepsilon_T\|_2$. Since $\EE[\|\bepsilon_T\|_2^2\mid\xb]
= d\sigma^2/D$, the same good-event risk decomposition as in Theorem~\ref{thm:single_manifold} yields
\begin{equation}
\label{eq:T1_union_bound}
T_1\le
4D\varepsilon^2
+4(1-q)^{-2}\frac{d\sigma^2}{D}
+8DB^2\PP(E_{q,*}^c).
\end{equation}

\noindent
\textbf{Step 4: Bounding the variance term.}
On $E_{q,*}$, we have $\|\yb\|_\infty<B+q\tau_*=:B_{q,*}$. Applying the same good-event covering-number argument as in Theorem~\ref{thm:single_manifold} on $[-B_{q,*},B_{q,*}]^D$ gives
\begin{equation}
\label{eq:T2_union_bound}
\begin{aligned}
T_2\le\;&
C(D^2\log^3D)n^{-1}\varepsilon^{-d}
\bigl(\log^4\varepsilon^{-1}
+\log^3\varepsilon^{-1}\log\eta^{-1}\bigr)\\
&+6\eta+4DB^2\PP(E_{q,*}^c).
\end{aligned}
\end{equation}

\noindent
\textbf{Step 5: Combining the bounds.}
Combining \eqref{eq:T1_union_bound} and \eqref{eq:T2_union_bound}, using the tail estimate in Lemma~\ref{lem:union_tubular_event}, and choosing $\varepsilon=n^{-1/(d+2)}$ and $\eta=n^{-1}$ as in Theorem~\ref{thm:single_manifold}, we obtain \eqref{eq:error_union}, completing the proof.
\end{proof}

\subsection{Proof of Theorem~\ref{thm:intersecting_union}}
\label{proof:intersecting_union}

\begin{proof}[Proof of Theorem~\ref{thm:intersecting_union}]
The proof follows the same five-step strategy as the proof of
Theorem~\ref{thm:union_manifold}, with an additional decomposition
according to whether the clean sample lies near the intersection set.

\noindent
\textbf{Step 1: Controlling the projection-valid and component-identification event.}

Define the non-intersection event $A_r:=\{\xb\notin\cI_r\}$, and the projection-valid and component-identification event $E_r:=\{\|\bepsilon\|_2<\rho_r\}$ with $\rho_r=\min\{q\tau_*,\Delta_r/2\}$.

On $A_r\cap E_r$, the clean sample lies in the non-intersection region.
Hence $\xb\in\cM_{\ell,r}$ for some component $\cM_\ell$. Since $\|\bepsilon\|_2<\rho_r\le q\tau_*$, the noisy observation remains within the projection-valid tubular neighborhood of the same component. Moreover, since $\rho_r\le\Delta_r/2$, the noisy observation remains separated from the other components. Thus, component identity is stable on
$A_r\cap E_r$.

\noindent
\textbf{Step 2: Error decomposition.}

We use the same population risk $L$, empirical risk $L_{\cS}$, and decomposition $ \EE_{\cS}L(\widehat{\cG}) = T_1+T_2$ as in \eqref{eq:risk_decomposition}.

\noindent
\textbf{Step 3: Bounding the bias term.}

We first extend the multi-chart oracle construction to the non-intersection region. Define the effective geometric scale $s_r:=\min\{\tau_*,\Delta_r\}$ and $q_r:=\frac{\rho_r}{s_r}$. By the definition of $\rho_r$, we have $0<q_r<1$. For each $\vb\in\cM_\ell$, define the truncated local scale $\bar{\tau}_{\ell,r}(\vb) =
\min\{\tau_{\cM_\ell}(\vb),\Delta_r\}$. Then $\bar{\tau}_{\ell,r}(\vb)\ge s_r$. Since both $\tau_{\cM_\ell}$ and the constant function $\Delta_r$ are $1$-Lipschitz, the truncated scale $\bar{\tau}_{\ell,r}$ is also $1$-Lipschitz.

For each component $\cM_\ell$, we repeat the projection-neighborhood partition-of-unity construction of \cite[Proposition~6.3]{cloninger2021deep}, using the relative radius
$q_r$ and the truncated local scale $\bar{\tau}_{\ell,r}$.
Since $\bar{\tau}_{\ell,r}(\vb) \le \tau_{\cM_\ell}(\vb)$, the resulting neighborhoods remain inside the projection-valid
region of $\cM_\ell$. The proof of \cite[Proposition~6.3]{cloninger2021deep} applies with this smaller geometric scale, after choosing the local separation parameter sufficiently small.

The truncation by $\Delta_r$ also prevents the local neighborhoods of the chart  from mixing different components. Indeed, suppose
$\xb\in\cM_{\ell,r}$ and $E_r$ holds. Then for any $\ell'\neq\ell$, $\operatorname{dist}(\yb,\cM_{\ell'})
\ge
2\Delta_r-\rho_r
\ge
\frac{3}{2}\Delta_r$. On the other hand, by the support property in \cite[Proposition~6.3]{cloninger2021deep}, if a local weight centered at $\vb\in\cM_{\ell'}$ is nonzero at $\yb$, then $\|\yb-\vb\|_2 < p\,\bar{\tau}_{\ell',r}(\vb) < \Delta_r$, since $p<1$ and $\bar{\tau}_{\ell',r}(\vb)\le\Delta_r$. This contradicts $ \operatorname{dist}(\yb,\cM_{\ell'}) \ge \frac{3}{2}\Delta_r$. Hence all local weights associated with components $\ell'\neq\ell$ vanish at $\yb$.

We then group the local partition functions into chart-level weights
exactly as in the proof of Lemma~\ref{lem:multichart_oracle}.
Writing these weights as $\rho_{\ell,j}$, we have, on $A_r\cap E_r$, for any $\ell'\neq\ell$, $\rho_{\ell',j}(\yb)=0$ and $\sum_{j=1}^{C_{\cM_\ell}}
\rho_{\ell,j}(\yb)=1$. Therefore, defining the multi-chart oracle encoder and decoder as in
Lemma~\ref{lem:multichart_oracle}, with the chart index extended to
$(\ell,j)$, gives
$$
\begin{aligned}
\cD_r^\circ\circ\cE_r^\circ(\yb)
&=
\sum_{j=1}^{C_{\cM_\ell}}
\phi_{\ell,j}^{-1}
\bigl(
\phi_{\ell,j}(\pi_\ell(\yb))
\bigr)
\rho_{\ell,j}(\yb)
=
\pi_\ell(\yb)
=
\pi(\yb),
\end{aligned}
$$
where the last equality follows from the stable component
identification on $A_r\cap E_r$.

Finally, the neural-network approximation follows the same
construction as in Lemma~\ref{lem:multichart_oracle_approx} and
Appendix~A.6 of \cite{liu2024chart}. The chart coordinates,
partition-of-unity functions, and inverse chart maps are approximated
componentwise and combined through the same multiplication and
summation construction. Since $s_r$ is a fixed geometric scale
independent of the approximation accuracy, this modification changes
only the constants in the network-size bounds and not their
asymptotic dependence on $\varepsilon$ and $D$. Thus, for any
$0<\varepsilon<1$, there exist $\widetilde{\cE}\in\cF_{\mathrm{NN}}^{\cE}$ and $\widetilde{\cD}\in\cF_{\mathrm{NN}}^{\cD}$ with the same asymptotic network-size scaling as in Theorem~\ref{thm:union_manifold}, such that, on $A_r\cap E_r$, writing $\widetilde{\cG}:=\widetilde{\cD}\circ\widetilde{\cE}$, we have
$$
\|\widetilde{\cG}(\yb)-\pi(\yb)\|_\infty
\le
\varepsilon
$$

As in the proof of Theorem~\ref{thm:single_manifold},
the global ERM property gives $T_1\le 2\inf_{\cG\in\cF_{\mathrm{NN}}^{\cG}}L(\cG)
\le 2L(\widetilde{\cG})$, where $\widetilde{\cG}$ is the neural approximation constructed above. We decompose the population risk
into three regions:
$$
\begin{aligned}
L(\widetilde{\cG})
=&\;
\EE\!\left[
\|\widetilde{\cG}(\yb)-\xb\|_2^2
\mathbf 1_{A_r\cap E_r}
\right]
+
\EE\!\left[
\|\widetilde{\cG}(\yb)-\xb\|_2^2
\mathbf 1_{A_r\cap E_r^c}
\right]
+
\EE\!\left[
\|\widetilde{\cG}(\yb)-\xb\|_2^2
\mathbf 1_{A_r^c}
\right].
\end{aligned}
$$
On $A_r\cap E_r$, the projection onto the correct component is uniquely
defined. Hence
$$
\|\widetilde{\cG}(\yb)-\xb\|_2^2
\le
2\|\widetilde{\cG}(\yb)-\pi(\yb)\|_2^2
+
2\|\pi(\yb)-\xb\|_2^2.
$$
By the oracle-approximation bound, $\|\widetilde{\cG}(\yb)-\pi(\yb)\|_2^2
\le
D\varepsilon^2$. Moreover, the same projection argument as in
Theorem~\ref{thm:union_manifold} gives
$
\|\pi(\yb)-\xb\|_2
\le
(1-q)^{-1}\|\bepsilon_T\|_2.
$
Therefore,
$$
\begin{aligned}
&\EE\!\left[
\|\widetilde{\cG}(\yb)-\xb\|_2^2
\mathbf 1_{A_r\cap E_r}
\right]
\le
2D\varepsilon^2
+
2(1-q)^{-2}
\EE\!\left[
\|\bepsilon_T\|_2^2
\mathbf 1_{A_r}
\right].
\end{aligned}
$$
Since $A_r$ depends only on $\xb$ and the Gaussian perturbation $\bepsilon$ is independent of $\xb$, while $
\EE\!\left[
\|\bepsilon_T\|_2^2
\mid \xb
\right]
=
\frac{d\sigma^2}{D}$, we obtain
$
\EE\!\left[
\|\bepsilon_T\|_2^2
\mathbf 1_{A_r}
\right]
=
(1-p_r)\frac{d\sigma^2}{D}.
$
Thus
$$
\begin{aligned}
&\EE\!\left[
\|\widetilde{\cG}(\yb)-\xb\|_2^2
\mathbf 1_{A_r\cap E_r}
\right]
\le
2D\varepsilon^2
+
2(1-q)^{-2}
(1-p_r)\frac{d\sigma^2}{D}.
\end{aligned}
$$
On $A_r\cap E_r^c$, boundedness gives $\|\widetilde{\cG}(\yb)-\xb\|_2^2
\le
4DB^2$. Since $A_r$ depends only on $\xb$ and $E_r$ depends only on $\bepsilon$, the two events are independent. Hence
$
\PP(A_r\cap E_r^c)
=
(1-p_r)\PP(E_r^c),
$
and therefore
$$
\EE\!\left[
\|\widetilde{\cG}(\yb)-\xb\|_2^2
\mathbf 1_{A_r\cap E_r^c}
\right]
\le
4DB^2(1-p_r)\PP(E_r^c).
$$
Finally, on $A_r^c$, boundedness gives $\EE\!\left[\|\widetilde{\cG}(\yb)-\xb\|_2^2 \mathbf 1_{A_r^c}\right]\le 4DB^2p_r$. Combining the three regions, using $T_1\le 2L(\widetilde{\cG})$, and applying the Gaussian tail estimate as in Lemma~\ref{lem:union_tubular_event} with $q\tau_*$ replaced by $\rho_r$, we obtain
\begin{equation}
\label{eq:T1_intersecting_bound_tail}
\begin{aligned}
\textstyle
T_1
\le\;&
4D\varepsilon^2
+
4(1-q)^{-2}
(1-p_r)\frac{d\sigma^2}{D}\\
&+
8DB^2(1-p_r)
\exp\!\left[
-\frac D4
\left(
\sqrt{\frac{2\rho_r^2}{\sigma^2}-1}-1
\right)^2
\right]
+
8DB^2p_r.
\end{aligned}
\end{equation}

\noindent
\textbf{Step 4: Bounding the variance term.}

On $E_r$, we have $\|\yb\|_\infty<B+\rho_r=:B_r$. Thus, the same good-event covering-number argument as in Theorem~\ref{thm:single_manifold} applies on
$[-B_r,B_r]^D$, yielding
\begin{equation}
\label{eq:T2_intersecting_bound}
\begin{aligned}
T_2\le\;&
C(D^2\log^3 D)n^{-1}\varepsilon^{-d}
\bigl(
\log^4\varepsilon^{-1}
+\log^3\varepsilon^{-1}\log\eta^{-1}
\bigr)\\
&+6\eta+4DB^2\PP(E_r^c).
\end{aligned}
\end{equation}
The remaining probability is controlled by the same
Gaussian norm tail bound used above.

\noindent
\textbf{Step 5: Combining the bounds.}

Combining \eqref{eq:T1_intersecting_bound_tail}
and \eqref{eq:T2_intersecting_bound},
using the Gaussian tail bound and
$8(1-p_r)+4\le12$, and choosing
$\varepsilon=n^{-1/(d+2)}$ and $\eta=n^{-1}$,
we obtain \eqref{eq:intersecting_union_bound},
completing the proof.

\end{proof}

\section{Conclusion}
\label{sec:conclusion}

In this paper, we studied the statistical performance of denoising autoencoders under the manifold hypothesis. We established generalization error bounds for three geometric settings: a single manifold, a finite union of pairwise separated manifolds, and a finite union of intersecting manifolds. In the first two settings, the sample-size-dependent statistical term scales as $n^{-\frac{2}{d+2}}\log^4 n$, up to its dependence on the ambient dimension, together with a tangential-noise term and a projection-failure tail term. For intersecting manifolds, we further identified an additional ambiguity contribution associated with the probability mass near the intersection set. Numerical experiments on synthetic manifolds and real image datasets support the predicted intrinsic-dimension-dependent sample-size behavior and illustrate the increased reconstruction ambiguity near manifold intersections.

\section*{Acknowledgment}

Haoran Yan and Wenjing Liao acknowledge support from the National Science Foundation under the NSF DMS 2145167 and the U.S. Department of Energy under the DOE SC0024348.  The initial manuscript, theoretical analysis, and experimental code were developed independently by the authors in early 2026. In Fall 2026, OpenAI ChatGPT was subsequently used to assist with manuscript review, mathematical verification, code refinement, and figure preparation. In particular, AI-assisted review helped identify two issues: the statistical error term in the intersecting-manifold analysis must be controlled using the full training sample rather than a conditional subset, and the covering-number argument requires localization to a bounded input domain under Gaussian noise. These issues were addressed in the revised proofs. AI tools also assisted with refining experimental designs, modifying analysis and plotting scripts, and regenerating selected figures. All mathematical arguments, code changes, and experimental results were reviewed and verified by the authors. The authors assume responsibility for all content.

\appendix

\section{Proofs of Auxiliary Lemmas}
\label{sec:appendix_lemmas}

\subsection{Proof of Lemma~\ref{lem:tubular_event}}
\label{app:tubular_event}

\begin{proof}[Proof of Lemma~\ref{lem:tubular_event}]
By the definition of $E_q$, $E_q^c = \left\{\|\bepsilon\|_2\ge q\tau_{\cM}\right\}$. Since $
\bepsilon
\sim
\mathcal N\!\left(
0,\frac{\sigma^2}{D}I_D
\right)$,
we have $\frac{D}{\sigma^2}\|\bepsilon\|_2^2
\sim
\chi_D^2$.
Therefore,
$
\PP(E_q^c)
=
\PP\!\left(
\chi_D^2
\ge
\frac{Dq^2\tau_{\cM}^2}{\sigma^2}
\right).
$
Set $a := \frac{q^2\tau_{\cM}^2}{\sigma^2}$. Since $q\tau_{\cM}>\sigma$, we have $a>1$.

Applying the Laurent--Massart inequality
\cite[Lemma~1]{laurent2000adaptive},
$
\PP\!\left(\chi_D^2-D \ge 2\sqrt{Dx}+2x
\right) \le e^{-x},
$
and choosing
$
x = \frac D4 \left(\sqrt{2a-1}-1 \right)^2,
$
gives $\PP(E_q^c) \le \exp\!\left[-\frac D4
\left(\sqrt{\frac{2q^2\tau_{\cM}^2}{\sigma^2}-1}-1 \right)^2 \right]$.
On the event $E_q$, $d(\yb,\cM) \le \|\yb-\xb\|_2 = \|\bepsilon\|_2 < q\tau_{\cM} < \tau_{\cM}$. Hence $\yb$ lies in the tubular neighborhood on which the
orthogonal projection $\pi(\yb)$ is uniquely defined.

Finally, conditioned on $\xb\in\cM$, the tangential component satisfies $\bepsilon_T \sim \mathcal N\!\left(
0,\frac{\sigma^2}{D}I_d \right)$, and therefore $\EE\!\left[\|\bepsilon_T\|_2^2 \mid \xb \right] = \frac{\sigma^2}{D}\EE[\chi_d^2] = \frac{d\sigma^2}{D}$.
\end{proof}

\subsection{Proof of Lemma~\ref{lem:multichart_oracle}}
\label{app:multichart_oracle}

\begin{proof}[Proof of Lemma~\ref{lem:multichart_oracle}]
We construct the oracle encoder and decoder in three steps.

\noindent
\textbf{Step 1: Construction of a local partition of unity on the projection neighborhood.}

Let $\{\vb_k'\}_{k=1}^c\subset\cM$ be a maximal
$\delta$-separated set with respect to the geodesic distance on $\cM$, and let $A_{\vb_k'}\in\RR^{D\times d}$ be the matrix whose columns form an
orthonormal basis of $T_{\vb_k'}\cM$. Following the construction in \cite[Proposition~6.3]{cloninger2021deep}, define $p:=(1+q)/2$ and $h:=6/(1-q/p)$. For sufficiently small $\delta>0$, with $\delta<C(1-q)^2\tau_{\cM}$ for a universal constant $C>0$, define
$$
\textstyle
\bar{\eta}_k(\yb)
=\max\Bigl\{
1-\frac{\|\yb-\vb_k'\|_2^2}{p^2\tau_{\cM}^2(\vb_k')}
-\frac{\|A_{\vb_k'}^\top(\yb-\vb_k')\|_2^2}{h^2\delta^2},
\,0
\Bigr\}.
$$
and normalize $\eta_k(\yb) = \frac{\bar{\eta}_k(\yb)}{\|\bar{\eta}(\yb)\|_1}$, where $\bar{\eta}(\yb) = [\bar{\eta}_1(\yb),\dots,\bar{\eta}_c(\yb)]^\top$. Then $\{\eta_k\}_{k=1}^c$ forms a partition of unity on the
projection-valid neighborhood.

\noindent
\textbf{Step 2: Grouping into finitely many chart domains.}

Following the multi-chart grouping construction in
\cite[Lemma~10]{liu2024chart}, there exist two atlases $\{(\widetilde V_j,\phi_j)\}_{j=1}^{C_{\cM}}$, $\{(\overline V_j,\phi_j)\}_{j=1}^{C_{\cM}}$, with $\overline V_j\subset\widetilde V_j$ for $j=1,\dots,C_{\cM}$ such that the chart maps satisfy the required uniform geometric
bounds. The number of charts $C_{\cM}$ is controlled by the covering
estimate in \cite[Lemma~12]{liu2024chart}.

We define the index sets by $ \mathcal I_j = \left\{
k:
(U_k\cap\cM)\cap\overline V_j\neq\varnothing
\ \text{and}\
k\notin\mathcal I_{j'}
\text{ for all }j'<j
\right\}
$, where $U_k:=\operatorname{supp}(\eta_k)$, and set $V_j := \bigcup_{k\in\mathcal I_j}(U_k\cap\cM)$, for $j=1,\dots,C_{\cM}$. This construction yields a finite atlas $\{(V_j,\phi_j)\}_{j=1}^{C_{\cM}}$ of $\cM$.

\noindent
{\bf Step 3: Definition of the oracle encoder and decoder.}

Define $\rho_j(\yb) := \sum_{k\in\mathcal I_j}\eta_k(\yb)$ for $j=1,\dots,C_{\cM}$. Then $\{\rho_j\}_{j=1}^{C_{\cM}}$ forms a partition of unity on
$\cM(q)$, and
$\{\rho_j|_{\cM}\}_{j=1}^{C_{\cM}}$ is subordinate to the atlas
$\{(V_j,\phi_j)\}_{j=1}^{C_{\cM}}$.
The chart maps and their inverses are understood as bounded Lipschitz extensions outside their original domains, following the extension argument in \cite[Lemma~7 and Appendix~A.6]{liu2024chart}. The oracle encoder is $\cE(\yb) = [\fb_1(\yb)\ \cdots\ \fb_{C_{\cM}}(\yb)]^\top$, where $\fb_j(\yb) =
\begin{bmatrix}
\phi_j(\pi(\yb))\\
\rho_j(\yb)
\end{bmatrix}
\in\RR^{d+1},$
and the oracle decoder is $\cD(\zb) = \sum_{j=1}^{C_{\cM}} \phi_j^{-1}\bigl((\zb_j)_{1:d}\bigr)\, (\zb_j)_{d+1}$.
For any $\yb\in\cM(q)$,$\cD\circ\cE(\yb) = \sum_{j=1}^{C_{\cM}} \phi_j^{-1}\bigl(\phi_j(\pi(\yb))\bigr)\, \rho_j(\yb) = \pi(\yb) \sum_{j=1}^{C_{\cM}}\rho_j(\yb) = \pi(\yb)$. This proves the claim.
\end{proof}

\subsection{Proof of Lemma~\ref{lem:multichart_oracle_approx}}
\label{app:multichart_oracle_approx}

\begin{proof}[Proof of Lemma~\ref{lem:multichart_oracle_approx}]
The proof follows the network construction in
\cite[Lemma~11 and Appendix~A.6]{liu2024chart}. For the encoder, the local chart coordinates $\phi_j\circ\pi$ and the partition-of-unity functions $\rho_j$
are approximated by ReLU networks and assembled across the $C_{\cM}$ charts to form
$\widetilde{\cE}\in\cF_{\mathrm{NN}}^{\cE}$.
For the decoder, the inverse chart maps $\phi_j^{-1}$ are approximated by ReLU networks, and the corresponding local reconstructions are combined with the partition weights to form $\widetilde{\cD}\in\cF_{\mathrm{NN}}^{\cD}$.

The construction in \cite[Appendix~A.6]{liu2024chart}
gives the stated bounds on
$L_{\cE},p_{\cE},K_{\cE},\kappa_{\cE},R_{\cE}$ and
$L_{\cD},p_{\cD},K_{\cD},\kappa_{\cD},R_{\cD}$, and yields
$
\sup_{\yb\in\cM(q)}
\left\|
\widetilde{\cD}\circ\widetilde{\cE}(\yb)
-
\cD\circ\cE(\yb)
\right\|_\infty
\le
\varepsilon.
$
By Lemma~\ref{lem:multichart_oracle}, for $\yb\in\cM(q)$, $\cD\circ\cE(\yb)=\pi(\yb)$. Therefore,
$
\sup_{\yb\in\cM(q)}
\left\|
\widetilde{\cD}\circ\widetilde{\cE}(\yb)
-
\pi(\yb)
\right\|_\infty
\le
\varepsilon.
$
\end{proof}

\subsection{Proof of Lemma~\ref{lem:variance_bound}}
\label{app:variance_bound}

\begin{proof}
Following the symmetrization argument in
\cite[Lemma~2]{liu2024chart}, we separate the good event
$E$ from its complement. Define $h_{\cG,E}(\yb,\xb):=\|\cG(\yb)-\xb\|_2^2\mathbf 1_E$ and $\cH_E:=\{h_{\cG,E}:\cG\in\cF_{\mathrm{NN}}^{\cG}\}$. Since $\xb,\cG(\yb)\in[-B,B]^D$, the squared loss is bounded by $4DB^2$. Dropping the nonpositive empirical contribution on $E^c$, we obtain
$$
\textstyle
T_2\le
\EE_{\cS}\left[
\EE h_{\widehat{\cG},E}
-\frac{2}{n}\sum_{i=1}^n
h_{\widehat{\cG},E}(\yb_i,\xb_i)
\right]
+4DB^2\PP(E^c).
$$
Applying the metric-entropy bound of \cite[Lemma~2]{liu2024chart} to the bounded loss class $\cH_E$ gives
$$
\textstyle
T_2\le
\frac{35DB^2}{n}
\log N(\eta,\cH_E,\|\cdot\|_\infty)
+6\eta+4DB^2\PP(E^c).
$$
For any $\cG,\cG'\in\cF_{\mathrm{NN}}^{\cG}$,
since $\yb\in[-B_E,B_E]^D$ on $E$, we have
$$
|h_{\cG,E}(\yb,\xb)-h_{\cG',E}(\yb,\xb)|
\le 4DB\,
\|\cG-\cG'\|_{L^{\infty,\infty}([-B_E,B_E]^D)}.
$$
Outside $E$, both truncated losses vanish. Hence
$$
\textstyle
N(\eta,\cH_E,\|\cdot\|_\infty)
\le
N\left(
\frac{\eta}{4DB},
\cF_{\mathrm{NN}}^{\cG},
\|\cdot\|_{L^{\infty,\infty}([-B_E,B_E]^D)}
\right).
$$
Substituting this estimate into the preceding bound completes the proof.
\end{proof}

\subsection{Proof of Lemma~\ref{lem:union_tubular_event}}
\label{app:union_projection}

\begin{proof}[Proof of Lemma~\ref{lem:union_tubular_event}]
The Gaussian tail bound follows from the same Laurent--Massart
argument as in Lemma~\ref{lem:tubular_event}, with $q\tau_{\cM}$ replaced by $q\tau_*$.

For the geometric claim, fix $\xb\in\cM_\ell$. On $E_{q,*}$, we have $\|\bepsilon\|_2<q\tau_*\le q\tau_{\cM_\ell}$, so the orthogonal projection of $\yb$ onto $\cM_\ell$ is uniquely defined. Moreover, for any $\ell'\neq\ell$, $d(\yb,\cM_{\ell'})
\ge 2\Delta-\|\bepsilon\|_2
>\Delta>\|\bepsilon\|_2
\ge d(\yb,\cM_\ell)$, where we used $q\tau_*<\Delta$.
Thus, $\cM_\ell$ is the unique closest component,
and the projection onto $\cM$ is uniquely defined
and belongs to $\cM_\ell$.
\end{proof}

\newpage
\bibliographystyle{siamplain}
\bibliography{ref}

@article{laurent2000adaptive,
  title={Adaptive estimation of a quadratic functional by model selection},
  author={Laurent, B. and Massart, P.},
  journal={The Annals of Statistics},
  volume={28},
  number={5},
  pages={1302--1338},
  year={2000}
}

@article{vincent2011connection,
  title={A Connection Between Score Matching and Denoising Autoencoders},
  author={Vincent, Pascal},
  journal={Neural Computation},
  volume={23},
  number={7},
  pages={1661--1674},
  year={2011}
}

@article{yang2023medmnist,
  title={MedMNIST v2-A large-scale lightweight benchmark for 2D and 3D biomedical image classification},
  author={Yang, Jiancheng and Shi, Rui and Wei, Donglai and Liu, Zequan and Zhao, Lin and Ke, Bilian and Pfister, Hanspeter and Ni, Bingbing},
  journal={Scientific Data},
  volume={10},
  number={1},
  pages={41},
  year={2023},
  publisher={Nature Publishing Group UK London}
}

@article{georghiades2001few,
  author={Georghiades, A. S. and Belhumeur, P. N. and Kriegman, D. J.},
  title={From Few to Many: Illumination Cone Models for Face Recognition under Variable Lighting and Pose},
  journal={IEEE Transactions on Pattern Analysis and Machine Intelligence},
  volume={23},
  number={6},
  pages={643--660},
  year={2001}
}

@article{lecun1998gradient,
  author={LeCun, Y. and Bottou, L. and Bengio, Y. and Haffner, P.},
  journal={Proceedings of the IEEE}, 
  title={Gradient-Based Learning Applied to Document Recognition}, 
  year={1998},
  volume={86},
  number={11},
  pages={2278--2324},
  doi={10.1109/5.726791}
}

@article{liu2024chart,
  title={Deep nonparametric estimation of intrinsic data structures by chart autoencoders: Generalization error and robustness},
  author={Liu, Hao and Havrilla, Alex and Lai, Rongjie and Liao, Wenjing},
  journal={Applied and Computational Harmonic Analysis},
  volume={68},
  pages={101602},
  year={2024},
  publisher={Elsevier}
}

@article{liu2022adaptive,
  title={Adaptive two-layer {ReLU} neural network: {I}. Best least-squares approximation},
  author={Liu, Min and Cai, Zhiqiang and Chen, Jingshuang},
  journal={Computers \& Mathematics with Applications},
  volume={113},
  pages={34--44},
  year={2022},
  publisher={Elsevier}
}

@article{havrilla2024understanding,
  title={Understanding scaling laws with statistical and approximation theory for transformer neural networks on intrinsically low-dimensional data},
  author={Havrilla, Alex and Liao, Wenjing},
  journal={Advances in Neural Information Processing Systems},
  volume={37},
  pages={42162--42210},
  year={2024}
}

@article{kramer1991nonlinear,
	author = {Kramer, Mark A},
	journal = {AIChE Journal},
	number = {2},
	pages = {233--243},
	publisher = {Wiley Online Library},
	title = {Nonlinear principal component analysis using autoassociative neural networks},
	volume = {37},
	year = {1991}}

@article{coifman2005geometric,
	author = {Coifman, Ronald R and Lafon, Stephane and Lee, Ann B and Maggioni, Mauro and Nadler, Boaz and Warner, Frederick and Zucker, Steven W},
	journal = {Proceedings of the National Academy of Sciences},
	number = {21},
	pages = {7426--7431},
	publisher = {National Acad Sciences},
	title = {Geometric diffusions as a tool for harmonic analysis and structure definition of data: {D}iffusion maps},
	volume = {102},
	year = {2005}}

@article{cloninger2021deep,
	author = {Cloninger, Alexander and Klock, Timo},
	journal = {Neural Networks},
	pages = {404--419},
	publisher = {Elsevier},
	title = {A deep network construction that adapts to intrinsic dimensionality beyond the domain},
	volume = {141},
	year = {2021}}

@article{chen2019nonparametric,
  title={Nonparametric regression on low-dimensional manifolds using deep ReLU networks: Function approximation and statistical recovery},
  author={Chen, Minshuo and Jiang, Haoming and Liao, Wenjing and Zhao, Tuo},
  journal={Information and Inference: A Journal of the IMA},
  volume={11},
  number={4},
  pages={1203--1253},
  year={2022},
  publisher={Oxford University Press}
}

@incollection{tu2011manifolds,
	author = {Tu, Loring W},
	booktitle = {An Introduction to Manifolds},
	pages = {47--83},
	publisher = {Springer},
	title = {Manifolds},
	year = {2011}}

@book{lee2006riemannian,
	author = {Lee, John M},
	publisher = {Springer Science \& Business Media},
	title = {Riemannian Manifolds: An Introduction to Curvature},
	volume = {176},
	year = {2006}}

@book{spivak1975comprehensive,
	author = {Spivak, Michael},
	publisher = {Publish or Perish, Incorporated},
	title = {A Comprehensive Introduction to Differential Geometry},
	volume = {4},
	year = {1975}}

@article{bourlard1988auto,
	author = {Bourlard, Herv{\'e} and Kamp, Yves},
	journal = {Biological Cybernetics},
	number = {4},
	pages = {291--294},
	publisher = {Springer},
	title = {Auto-association by multilayer perceptrons and singular value decomposition},
	volume = {59},
	year = {1988}}

@article{hinton1993autoencoders,
	author = {Hinton, Geoffrey E and Zemel, Richard},
	journal = {Advances in Neural Information Processing Systems},
	title = {Autoencoders, minimum description length and {H}elmholtz free energy},
	volume = {6},
	year = {1993}}

@article{liou2014autoencoder,
	author = {Liou, Cheng-Yuan and Cheng, Wei-Chen and Liou, Jiun-Wei and Liou, Daw-Ran},
	journal = {Neurocomputing},
	pages = {84--96},
	publisher = {Elsevier},
	title = {Autoencoder for words},
	volume = {139},
	year = {2014}}

@article{fefferman2016testing,
	author = {Fefferman, Charles and Mitter, Sanjoy and Narayanan, Hariharan},
	journal = {Journal of the American Mathematical Society},
	number = {4},
	pages = {983--1049},
	title = {Testing the manifold hypothesis},
	volume = {29},
	year = {2016}}

@article{tenenbaum290global,
	author = {Tenenbaum, Joshua B and Silva, Vin de and Langford, John C},
	journal = {Science},
	number = {5500},
	pages = {2319--2323},
	publisher = {American Association for the Advancement of Science},
	title = {A global geometric framework for nonlinear dimensionality reduction},
	volume = {290},
	year = {2000}}

@article{roweis2000nonlinear,
	author = {Roweis, Sam T and Saul, Lawrence K},
	journal = {Science},
	number = {5500},
	pages = {2323--2326},
	publisher = {American Association for the Advancement of Science},
	title = {Nonlinear dimensionality reduction by locally linear embedding},
	volume = {290},
	year = {2000}}

@article{shaham2018provable,
	author = {Shaham, Uri and Cloninger, Alexander and Coifman, Ronald R},
	journal = {Applied and Computational Harmonic Analysis},
	number = {3},
	pages = {537--557},
	publisher = {Elsevier},
	title = {Provable approximation properties for deep neural networks},
	volume = {44},
	year = {2018}}

@article{chen2019efficient,
	author = {Chen, Minshuo and Jiang, Haoming and Liao, Wenjing and Zhao, Tuo},
	journal = {Advances in Neural Information Processing Systems},
	title = {Efficient approximation of deep {ReLU} networks for functions on low dimensional manifolds},
	volume = {32},
	year = {2019}}

@article{nakada2020adaptive,
	author = {Nakada, Ryumei and Imaizumi, Masaaki},
	journal = {J. Mach. Learn. Res.},
	number = {174},
	pages = {1--38},
	title = {Adaptive Approximation and Generalization of Deep Neural Network with Intrinsic Dimensionality.},
	volume = {21},
	year = {2020}}

@article{liu2022benefits,
	author = {Liu, Hao and Chen, Minshuo and Er, Siawpeng and Liao, Wenjing and Zhang, Tong and Zhao, Tuo},
	journal = {arXiv preprint arXiv:2206.04569},
	title = {Benefits of Overparameterized Convolutional Residual Networks: {F}unction Approximation under Smoothness Constraint},
	year = {2022}}

@article{schonsheck2019chart,
	author = {Schonsheck, Stefan and Chen, Jie and Lai, Rongjie},
	journal = {arXiv preprint arXiv:1912.10094},
	title = {Chart auto-encoders for manifold structured data},
	year = {2019}}

@article{thale200850,
	author = {Th{\"a}le, Christoph},
	journal = {Surveys in Mathematics and its Applications},
	pages = {123--165},
	publisher = {University Constantin Brancusi},
	title = {50 years sets with positive reach--a survey.},
	volume = {3},
	year = {2008}}

@inproceedings{pope2020intrinsic,
	author = {Pope, Phil and Zhu, Chen and Abdelkader, Ahmed and Goldblum, Micah and Goldstein, Tom},
	booktitle = {International Conference on Learning Representations},
	title = {The Intrinsic Dimension of Images and Its Impact on Learning},
	year = {2020}}

@inproceedings{vincent2008extracting,
	author = {Vincent, Pascal and Larochelle, Hugo and Bengio, Yoshua and Manzagol, Pierre-Antoine},
	booktitle = {Proceedings of the 25th International Conference on Machine Learning},
	pages = {1096--1103},
	title = {Extracting and composing robust features with denoising autoencoders},
	year = {2008}}

@article{bengio2009learning,
	author = {Bengio, Yoshua and others},
	journal = {Foundations and Trends{\textregistered} in Machine Learning},
	number = {1},
	pages = {1--127},
	publisher = {Now Publishers, Inc.},
	title = {Learning deep architectures for {AI}},
	volume = {2},
	year = {2009}}

@article{federer1959curvature,
	author = {Federer, Herbert},
	journal = {Transactions of the American Mathematical Society},
	number = {3},
	pages = {418--491},
	publisher = {JSTOR},
	title = {Curvature measures},
	volume = {93},
	year = {1959}}

@article{niyogi2008finding,
	author = {Niyogi, Partha and Smale, Stephen and Weinberger, Shmuel},
	journal = {Discrete \& Computational Geometry},
	number = {1-3},
	pages = {419--441},
	publisher = {Springer},
	title = {Finding the homology of submanifolds with high confidence from random samples},
	volume = {39},
	year = {2008}}

\end{document}